\newif\ifelsevier
\elsevierfalse

\newif\ifojies
\ojiestrue

\newif\ifieeet
\ieeetfalse

\ifelsevier
    \documentclass[review]{elsarticle}
    \usepackage{lineno,hyperref}
    \modulolinenumbers[5]

    \journal{Neural Networks}

\biboptions{authoryear}

    \usepackage[title]{appendix}
\fi

\ifojies
    \documentclass{ieeeojiespre}
    \usepackage{cite}
    \usepackage{amsmath,amssymb,amsfonts}
    \usepackage{mathtools}
    \usepackage{algorithmic}
    \usepackage{graphicx}
    \usepackage{textcomp}
        
    \newcommand{\citep}[1]{\cite{ #1 }}
    \usepackage{multirow} 
    \usepackage{adjustbox}
    \usepackage{array}
    \usepackage{booktabs}
    \newcolumntype{R}[2]{%
        >{\adjustbox{angle=#1,lap=\width-(#2)}\bgroup}%
        l%
        <{\egroup}%
    }
    
    \def\BibTeX{{\rm B\kern-.05em{\sc i\kern-.025em b}\kern-.08em
        T\kern-.1667em\lower.7ex\hbox{E}\kern-.125emX}}
\fi
    
\ifieeet
    \documentclass[conference]{IEEEtran}

    \usepackage{cite}
    \usepackage{mathtools}
    \newcommand{\citep}[1]{\cite{ #1 }}

    \usepackage{multirow} 
    \usepackage{adjustbox}
    \usepackage{array}
    \usepackage{booktabs}
    \newcolumntype{R}[2]{%
        >{\adjustbox{angle=#1,lap=\width-(#2)}\bgroup}%
        l%
        <{\egroup}%
    }
\fi

\usepackage{url}            
\ifelsevier
    \usepackage{booktabs}       
    \usepackage{adjustbox}      
\fi

\usepackage{amsfonts}       
\usepackage{nicefrac}       
\usepackage{microtype}      

\usepackage{graphicx}
\usepackage{subfig}
\usepackage{enumerate} 

\usepackage{wrapfig}

\usepackage{colortbl}
  
\usepackage{color,soul}
\soulregister\cite7
\soulregister\ref7

\DeclareMathOperator*{\argmax}{arg\,max}

\begin{document}

\newcommand{\mat}[1]{ \begin{bmatrix} #1 \end{bmatrix} }
\newcommand{\ordered}[2]{ \D_{#1}^{\ct}(#2) }
\newcommand{\normal}[1]{ {\cal N} \left( #1 \right) }
\newcommand{\group}[1]{ \left( #1 \right) }
\newcommand{\sgroup}[1]{ \left[ #1 \right] }
\newcommand{\vect}[1]{ {\boldsymbol{#1}} }
\newcommand{\tensor}[1]{ {\textbf{\underline{#1}}} }
\newcommand{\diag}{ {\text{diag}} }
\newcommand{\vdiag}{ {\textbf{diag}} }
\newcommand{\ltwogroup}[1]{\left\lVert #1 \right\rVert}
\newcommand{\fnorm}[1]{\left\lVert #1 \right\rVert_F}
\newcommand{\abs}[1]{\left\lvert #1 \right\rvert}
\newcommand{\func}[2]{ { #1 \group{#2} } }
\newcommand{\dataset}{{\cal D}}
\newcommand{\eye}{ \vect{I}}
\newcommand{\va}{\vect{a}}
\newcommand{\vA}{\vect{A}}
\newcommand{\tA}{\tensor{A}}
\newcommand{\vb}{\vect{b}}
\newcommand{\vB}{\vect{B}}
\newcommand{\tB}{\tensor{B}}
\newcommand{\vc}{\vect{c}}
\newcommand{\vC}{\vect{C}}
\newcommand{\tC}{\tensor{C}}
\newcommand{\vf}{\vect{f}}
\newcommand{\vF}{\vect{F}}
\newcommand{\tF}{\tensor{F}}
\newcommand{\vg}{\vect{g}}
\newcommand{\vG}{\vect{G}}
\newcommand{\tG}{\tensor{G}}
\newcommand{\vh}{\vect{h}}
\newcommand{\vH}{\vect{H}}
\newcommand{\tH}{\tensor{H}}
\newcommand{\vk}{\vect{k}}
\newcommand{\vK}{\vect{K}}
\newcommand{\tK}{\tensor{K}}
\newcommand{\vm}{\vect{m}}
\newcommand{\vM}{\vect{M}}
\newcommand{\tM}{\tensor{M}}
\newcommand{\vp}{\vect{p}}
\newcommand{\vP}{\vect{P}}
\newcommand{\tP}{\tensor{P}}
\newcommand{\vq}{\vect{q}}
\newcommand{\vQ}{\vect{Q}}
\newcommand{\tQ}{\tensor{Q}}
\newcommand{\vr}{\vect{r}}
\newcommand{\vR}{\vect{R}}
\newcommand{\tR}{\tensor{R}}
\newcommand{\vs}{\vect{s}}
\newcommand{\vS}{\vect{S}}
\newcommand{\tS}{\tensor{S}}
\newcommand{\vu}{\vect{u}}
\newcommand{\vU}{\vect{U}}
\newcommand{\tU}{\tensor{U}}
\newcommand{\vw}{\vect{w}}
\newcommand{\vW}{\vect{W}}
\newcommand{\tW}{\tensor{W}}
\newcommand{\vx}{\vect{x}}
\newcommand{\vX}{\vect{X}}
\newcommand{\tX}{\tensor{X}}
\newcommand{\vy}{\vect{y}}
\newcommand{\vY}{\vect{Y}}
\newcommand{\tY}{\tensor{Y}}
\newcommand{\vz}{\vect{z}}
\newcommand{\vZ}{\vect{Z}}
\newcommand{\tZ}{\tensor{Z}}
\newcommand{\vbeta}{\vect{\beta}}
\newcommand{\vBeta}{\vect{\Beta}}
\newcommand{\tBeta}{\tensor{\Beta}}
\newcommand{\vmu}{\vect{\mu}}
\newcommand{\vsigma}{\vect{\sigma}}
\newcommand{\vSigma}{\vect{\Sigma}}
\newcommand{\vtau}{\vect{\tau}}
\newcommand{\vepsilon}{\vect{\epsilon}}
\newcommand{\vdelta}{\vect{\delta}}
\newcommand{\vDelta}{\vect{\Delta}}
\newcommand{\vLambda}{\vect{\Lambda}}
\newcommand{\D}{{\cal D}}
\newcommand{\ct}{{\dagger}}
\newcommand{\ones}{{\textbf{\underline{1}}}}
\newcommand{\vones}{{\textbf{1}}}
\newcommand{\defeq}{{\stackrel{def}{=}}}
\newcommand{\fracpartial}[2]{\frac{\partial #1}{\partial  #2}}
\newcommand{\grad}{\nabla}
\newcommand{\loss}{\mathcal{L}}
\newcommand{\set}[1]{{\left\{#1\right\}}}
\newcommand{\squareb}[1]{{\left[#1\right]}}
\newcommand{\lb}{\left(}
\newcommand{\rb}{\right)}
\newcommand{\reals}{{\mathbb R}}
\newcommand{\pd}{{\mathbb S}}

\newcommand{\Normal}{{\cal N}}
\newcommand{\uniform}[1]{\func{{\cal U}}{#1}}
\newcommand{\expected}[2]{ \mathop{{\mathbb E}}_{#1} \sgroup{#2} }
\newcommand{\expectedf}[2]{ \mathop{{\mathbb E}}_{#1} #2 }
\newcommand{\detm}[1]{\lvert #1 \rvert}
\newcommand{\Tr}[1]{\text{Tr}\group{#1}}
\newcommand{\GP}[1]{ {\cal GP} \group{ #1 } }

\newcommand{\deltat}{\Delta{t}}

\newcommand{\delequal}{\overset{\Delta}{=}}

\newcommand{\datasize}{{\abs{\dataset}}}

\ifelsevier
    \begin{frontmatter}
\fi

\title{Physics Enhanced Data-Driven Models with Variational Gaussian Processes}

\ifojies
    \author{
        \uppercase{Daniel L. Marino}\authorrefmark{1},
        \uppercase{Milos Manic}\authorrefmark{2},
        \IEEEmembership{Member, IEEE} 
    }
    \address[1]{Department of Computer Science, Virginia Commonwealth University, Richmond, VA 23284 USA (e-mail: marinodl@vcu.edu)}
    \address[2]{Department of Computer Science, Virginia Commonwealth University, Richmond, VA 23284 USA (e-mail: misko@ieee.org)}

    \markboth
    {Marino \headeretal: Physics Enhanced Data-Driven Models with Variational Gaussian Processes}
    {Marino \headeretal: Physics Enhanced Data-Driven Models with Variational Gaussian Processes}

    \corresp{Corresponding author: Daniel L. Marino (e-mail: marinodl@vcu.edu).}
\fi

\begin{abstract}

Centuries of development in natural sciences and mathematical modeling provide valuable domain expert knowledge that has yet to be explored for the development of machine learning models.
When modeling complex physical systems, both domain knowledge and data provide necessary information about the system.
In this paper, we present a data-driven model that takes advantage of partial domain knowledge in order to improve generalization and interpretability.
The presented approach, which we call EVGP (Explicit Variational Gaussian Process), has the following advantages:
1) using available domain knowledge to improve the assumptions (inductive bias) of the model, 2) scalability to large datasets, 3) improved interpretability.
We show how the EVGP model can be used to learn system dynamics using basic Newtonian mechanics as prior knowledge.
We demonstrate how the addition of prior domain-knowledge to data-driven models outperforms purely data-driven models.

\end{abstract}

\ifelsevier
    \end{frontmatter}
\fi
\ifojies
    \titlepgskip=-15pt
    \maketitle
\fi

    \section{Introduction}\label{introduction}

For centuries, scientists and engineers have worked on creating
  mathematical abstractions of real world systems.
This principled modeling approach provides a powerful toolbox to derive white-box
  models that we can use to understand and analyze physical systems.
However, as the complexity of physical systems grow, deriving detailed principled models
  becomes an expensive and tedious task that requires highly experienced scientists
  and engineers.
Moreover, incorrect assumptions leads to inaccurate models that are unable to
  represent the real system.

Data-driven black-box models provide an appealing alternative modeling approach
  that requires little to none domain knowledge.
These models are fit to data extracted from the real system, minimizing the
  problems derived from incorrect assumptions.
However, using data-driven models while
  completely ignoring domain knowledge may lead to models that do not generalize
  well and are hard to understand.
Completely black-box approaches ignore the structure of the problem,
  wasting resources \citep{szegedy2015going} and making the model less
  explainable \citep{adadi2018peeking}.

Gray-box models combine domain knowledge and data as both
	provide important and complementary information about the system.
Domain knowledge can be used to construct a set of basic assumptions about the
	system, giving the data-driven model a baseline to build upon.
Data can be used to fill the gaps in knowledge and model complex relations
  that were not considered by the domain expert.

In this paper, we explore an approach for embedding domain knowledge into a
  data-driven model in order to improve generalization and interpretability.
The presented gray-box model, which we called EVGP (Explicit Variational Gaussian Process),
  is a scalable approximation of a sparse Gaussian Process (GP) that uses
  domain knowledge to define the prior distribution of the GP.
In this paper domain knowledge is extracted from physics-based knowledge, however the
  EVGP can be applied to any domain.

\begin{figure}
  \centering
  \begin{minipage}{.5\textwidth}
    \centering
    \includegraphics[scale=0.6]{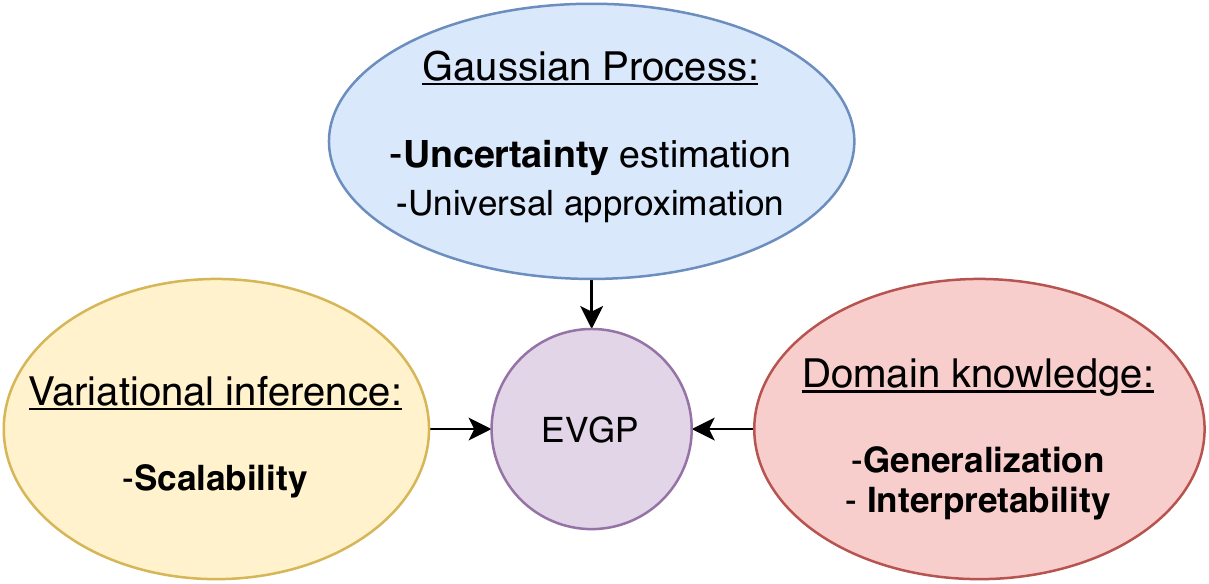}
    \caption{EVGP: Variational Gaussian-Process with explicit features}
    \label{figure:overview}
  \end{minipage}%
\end{figure}

The work on this paper has three cornerstones (Fig. \ref{figure:overview}):
  1) Gaussian processes are used for learning complex non-linear behavior from data
  	and model uncertainty,
  2) Partial domain knowledge is used as prior in order to improve inductive bias,
  3) Variational Inference provides advantageous scalability to large datasets.
Inductive bias refers to the assumptions made by the model when doing predictions
  over inputs that have not been observed.
The presented approach provides uncertainty estimations which are fundamental
  in order to avoid the risk associated with overconfidence in unexplored areas
  \citep{frigola2014variational} and warns the user of
  possible incorrect estimations \citep{marino2017data}.

The work in this paper is highly applicable when:
  1) modeling physical systems with uncertainty estimations,
  2) partial domain knowledge of the system is available,
  3) large quantities of data are available.
The aim is to help the engineer and take advantage of available knowledge without
  requiring the derivation of complex and detailed models.
Instead, an engineer only has to provide simple, partially formed, models and
  the EVGP takes care of filling the gaps in knowledge.
We show how the EVGP model can be used to learn system dynamics using
  basic physics laws as prior knowledge.

The \textbf{contributions} of this paper are:
\begin{itemize}
    \item We present a stochastic data-driven model enhanced with domain knowledge to improve model accuracy using less amount of data when compared with fully data-driven approaches.
    \item The presented approach uses very simple priors, demonstrating that accuracy gets improved by incorporating priors derived from heavily simplified physics.
    \item Scalability to handle large datasets is achieved by using the sparse variational framework.
    \item Interpretability is exemplified using the posterior distribution of trainable parameters, i.e. the value of the model parameters after training.
\end{itemize}

Our \textbf{objective} is to evaluate the benefits of including partially defined (imperfect) models as domain-knowledge to data-driven models.
As such, our experiments compare the performance of the EVGP approach with respect to fully data-driven (black-box) models to evaluate the benefits of the proposed approach.

The rest of the paper is organized as follows: 
section \ref{section:evgp} presents the proposed EVGP approach;
section \ref{section:physics-evgp} presents a set of priors derived from simplified Newtonian dynamics for the EVGP model;
section \ref{section:experiments} presents the experimental section where we compare the EVGP with fully data-driven models;
section \ref{section:related-work} presents the related work;
section \ref{section:conclusion} concludes the paper.

\newcommand{\vtestX}{\hat{\vX}}
\newcommand{\vtestx}{\hat{\vx}}
\newcommand{\vtestg}{\hat{\vg}}
\newcommand{\testx}{\hat{x}}
\newcommand{\testf}{\hat{\vf}}
\newcommand{\testy}{\hat{\vy}}
\newcommand{\vtesty}{\hat{\vy}}
\newcommand{\param}{\omega}
\newcommand{\linw}{\vbeta}
\newcommand{\accel}[1]{\ddot{ #1 }}
\newcommand{\vel}[1]{\dot{ #1 }}

\section{EVGP approach - variational GP using explicit features}
  \label{section:evgp}

\begin{figure} 
    \centering
    \includegraphics[scale=0.75]{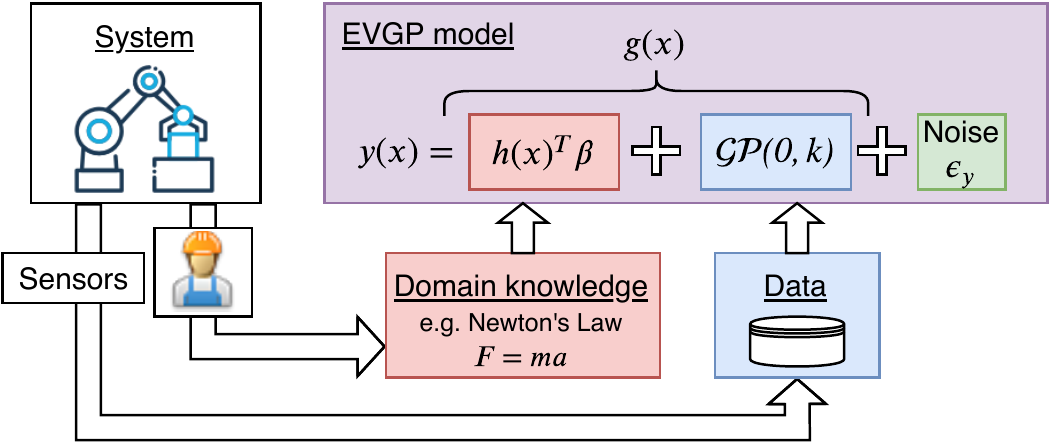}
    \caption{Illustration of the EVGP model. }
    \label{figure:model_intuition}
\end{figure}

  The novel EVGP approach presented in this paper is designed to solve regression problems under uncertainty. Figure \ref{figure:model_intuition} offers a visual representation of the approach.
  
  Given a dataset \(\dataset = (\vx^{(i)} , \vy^{(i)} ) \) composed
    of input/output pairs of samples $ \group{\vx^{(i)} , \vy^{(i)} }$,
    we would like to obtain a predictive distribution $p(\vy|\vx, \dataset)$
    that estimates the value of the output $\vy$ for a given input $\vx$.
  The EVGP model approximates $p(\vy|\vx, \dataset)$ using variational Inference.
  The EVGP is defined as a distribution $p(\vy|\vx, w)$ where $w$ are a set of
    parameters with prior distribution $p(w)$.

  In the following sections we describe in detail:
    \textit{A)} the EVGP model approach,
    \textit{B)} the variational loss function used to train the model,
    \textit{C)} the predictive distribution that approximates $p(\vy|\vx, \dataset)$.

\subsection{EVGP Model Definition}\label{model-definition}
  \label{section:model_definition}
  The EVGP model takes the following form:
  \begin{align}
    y = g(\vx) + \epsilon_y ; \;\;\;
    \func{g}{\vx} = \func{h}{\vx}^T \linw + \func{f}{\vx} \label{eq:evgp}
  \end{align}
where
  $\func{f}{\vx} \sim \GP{0, \func{k}{\vx, \vx'}}$ is a Gaussian process
    with kernel $k$,
  $\epsilon_y \sim \normal{0, \vSigma_y}$ is the observation noise and
  $g(\vx)$ is the denoised prediction for the input $\vx$.
Figure \ref{figure:model_intuition} offers a visual representation of the model.
The following is the description of the main components of the EVGP model:
\begin{itemize}
  \item Domain knowledge is embedded in the explicit function $\func{h}{\vx}^T \linw$,
    parameterized by $\linw$. The function $\func{h}{\vx}$ describes a set of
    explicit features (hence the name of our method) provided by the domain expert.
    $\linw$ is modeled using a normal distribution with a prior that is also extracted
      from domain knowledge.
    In this paper, $h(\vx)^T \linw$ is derived from partially defined Newtonian mechanics.

  \item The Gaussian Process $f(\vx)$ adds the ability to learn complex
      non-linear relations that $\func{h}{\vx}^T \linw$ is unable to
      capture.
\end{itemize}

Given a dataset $\dataset$, the exact predictive distribution $p(y|x, \dataset)$
  for the model in Eq. \refeq{eq:evgp} is described in \citep{rasmussen2004gaussian}.
For the rest of the paper, we refer to the exact distribution as EGP.
Computing the EGP predictive distribution has a large
  computational cost and does not scale well for large datasets.
To alleviate this problem, sparse approximation methods \citep{quinonero2005unifying}
	use a small set of $m$ inducing points $(\vf_m, \vX_m)$ (trainable parameters)
	instead of the entire dataset to approximate the predictive distribution.

In order to construct a sparse approximation for the model in Eq. \refeq{eq:evgp},
  we use a set of $m$ inducing points \((\vf_m, \vX_m)\) as parameters that
  will be learned from data.
Given  \((\vf_m, \vX_m)\) and a set of
  test points \((\vg, \vX)\), the prior distribution of the model can be expressed as follows:
\begin{align*}
\mat{\vg_m \mid \linw \\
     \;\; \vg \mid \linw }
    \sim
    \normal{\mat{\vH_m \linw \\
                 \vH_x, \linw},
            \mat{\vK_{m m}   & \vK_{m \vx} \\
                 \vK_{\vx m} & \vK_{\vx \vx}}};
\end{align*}
\begin{align*}
      \vH_m  &= \func{h}{\vX_m} ; \;\;\;
      \vH_x  = \func{h}{\vX} \\
      \vg_m &= h(\vX_m)\linw + \vf_m
\end{align*}
where $\vX$ denotes the data matrix, where each row represents an individual sample.
The rows of $\vH_x$ represent the value of the function $h()$
  applied to the real samples $\vX$.
The rows of $\vH_m$ represent the value of the function $h()$
  applied to the inducing (learned) points $\vX_m$.
Using the conditional rule for multivariate Gaussian distributions,
 we obtain the definition of the denoised sparse EVGP model:
\begin{align}
    \func{p}{\vg \mid \vX, \param} \sim &
      \normal{\vH_x \linw + \vmu_{\vf \mid \param},
              \vSigma_{\vf \mid \param}} \label{eq:g_given_w}
\end{align}
where $\param = \set{\vf_m, \linw}$ are the parameters of our model,
  $\vmu_{\vf \mid \param}
         = \vK_{\vx m}\vK_{mm}^{-1}\vf_m$, and
  $\vSigma_{\vf \mid \param}
         = \vK_{\vx \vx} - \vK_{\vx m} \vK_{mm}^{-1} \vK_{mx}$.
Equation \refeq{eq:g_given_w} defines our scalable EVGP model.
In following sections we give prior distributions to the parameters $\param$
	and perform approximate Bayesian inference.
Note that Eq. \refeq{eq:g_given_w} is also conditioned on $\vX_m$,
  however we do not indicate this explicitly in order to improve readability.

\subsection{Variational Loss}\label{variational-lower-bound}
In this section we present the loss function that we use to fit our model.
In this paper we follow a variational Bayesian approach
  (see Appendix \ref{section:variational_inference} for a brief overview).
Given a training dataset $\dataset$ and a prior distribution $p(\param)$,
  we wish to approximate the posterior distribution $p(\param|\dataset)$.
The posterior of $\param$ is
  approximated using a variational distribution $\func{p_\phi}{w} \approx p(\param|\dataset)$
  parameterized by $\phi$.
For the EVGP parameters $\param = \set{\vf_m, \linw}$,
  we use the following variational posterior distributions:
\begin{align*}
    \func{p_\phi}{\vf_m} = \normal{\vf_m | \va, \vA} ; \;\;\;
    \func{p_\phi}{\linw} = \normal{\linw | \vb, \vB}
\end{align*}

The prior-distributions for $\param$ are also defined as multivariate normal distributions:
\begin{align*}
    \func{p}{\vf_m} = \normal{\vf_m \mid 0, \vK_{mm}} ; \;\;\;
    \func{p}{\linw} = \normal{\linw \mid \vmu_\beta, \vSigma_{\beta}}
\end{align*}
these prior distributions represent our prior knowledge, i.e.
  our knowledge before looking at the data.

Given the training dataset $\dataset = (\vy, \vX)$, the parameters $\phi$
  of $\func{p_\phi}{\param}$ are learned by minimizing the negative
  Evidence Lower Bound (ELBO). For the EVGP, the negative ELBO takes the following
  form:
\begin{align}
  \func{\loss}{\phi}
  = & - \log \normal{\vy \mid \vH_x \vb + \vK_{xm} \vK_{mm}^{-1} \va, \vSigma_y}
            \nonumber \\
          & + \dfrac{1}{2} \sgroup{
               \Tr{\vM_1 \vA}
               + \Tr{\vM_2 \vB}
               + \Tr{\vSigma_y^{-1} \vSigma_{\vf \mid \param}}
               }  
            + \loss_{KL}  
          \label{eq:elbo}
\end{align}
where
$\vM_1 = \group{\vK_{mm}^{-1} \vK_{mx}} \vSigma_y^{-1} \group{\vK_{xm} \vK_{mm}^{-1}}$,
and $\vM_2 = \vH_x^T \vSigma_y^{-1} \vH_x$.
The term $\loss_{KL}$ is the KL-divergence between the posterior and prior
  distributions for the parameters:
  \begin{align*}
    \loss_{KL} = &
        D_{KL}\group{ \normal{\va, \vA} \mid \mid \normal{0, \vK_{mm}}} \\
         & + D_{KL}\group{ \normal{\vb, \vB} \mid\mid \normal{\vmu_{\beta}, \vSigma_{\beta}} }
  \end{align*}

A detailed derivation of the variational loss in Eq. \refeq{eq:elbo} is presented
  in Appendix \ref{appendix:elbo}.
The negative ELBO (Eq. \ref{eq:elbo}) serves as our loss function to learn the
  parameters $\phi$ given the training dataset $\vy, \vX$.
In order to scale to very large datasets, the ELBO is optimized using mini-batches
  (see Appendix \ref{appendix:minibatches}).
In our case, the parameters of the variational approximation are:
  $\phi={\va, \vA, \vb, \vB, \vX_m}$.

    \subsection{Predictive distribution}\label{section:evgp-prediction}
  After learning the parameters $\phi$, we would like to provide
    estimations using the approximated variational distribution $\func{p_\phi}{\param}$.
  Given a set of test points $\vtestX$, the estimated denoised predictive distribution
  	is computed as an expectation of Eq. \refeq{eq:g_given_w} w.r.t. $\func{p_\phi}{\param}$:

    \begin{align}
    \func{p_\phi}{\vtestg \mid \vtestX}
        =& \expected{p_\phi(\param)}{\func{p}{\vtestg \mid \vtestX, \param}} 
        = \normal{\vtestg \middle\vert \vmu_{\vtestg|\vtestx}, \vSigma_{\vtestg|\vtestx}} \label{eq:prediction} \\
      \vmu_{\vtestg|\vtestx} =& \vH_x \vb + \vK_{\testx m} \vK_{mm}^{-1} \va  \nonumber \\
      \vSigma_{\vtestg|\vtestx} =&
      \vSigma_{\vf \mid \param}
        + \vH_{\testx} \vB \vH_{\testx}^T 
        + \vK_{\testx m}\vK_{mm}^{-1} \vA \vK_{mm}^{-1} \vK_{m \testx} \nonumber
    \end{align}

Note that $\vtestg$ is just a denoised version of $\vtesty$.
Eq. \refeq{eq:prediction} approximates $p(\vtestg | \vtestx, \dataset)$,
  using the learned distribution $\func{p_\phi}{\param}$ (see Appendix \ref{appendix:elbo}).
The result is equivalent to \citep{titsias2009variational} with the addition of
  $\vH_x \vb$ for the mean and $\vH_{\testx} \vB \vH_{\testx}^T$ for the covariance.
These additional terms include the information provided by the prior function $\vH_x$
  with the parameters $\vb$ and $\vB$ that were learned from data.

  The approximated predictive distribution with observation noise is the following:
  \begin{align}
    \func{p_\phi}{\vtesty \mid \vtestX} =
      \normal{\vtestg \middle\vert \vmu_{\vtestg|\vtestx},
              \vSigma_{\vtestg|\vtestx} + \vSigma_y} \nonumber
  \end{align}
  where $\func{p_\phi}{\vtesty \mid \vtestX} \approx p(\vtesty \mid \vtestx, \dataset)$.
  In the next section, we show how Eq. \refeq{eq:prediction} can be used
    to model system dynamics and predict the next state of a physical system
    given the control input and current state.

\section{Embedding physics-based knowledge}
  \label{section:physics-evgp}
\newcommand{\vstate}{\vz}
\newcommand{\vcontrol}{\vu}
\newcommand{\vsensor}{\vy}

In this paper, we apply the EVGP model to learn the dynamics of a physical system.
The state $\vstate_{[t]}$ of the physical system can be modeled as follows:
\begin{align}
  \vstate_{[t+1]} &\sim g(\vz_{[t]} \oplus \vu_{[t]}) \label{eq:dynamic_evgp}\\
  \vsensor_{[t]}   &\sim \vz_{[t]} + \epsilon_y \nonumber
\end{align}
where $\vcontrol_{[t]}$ is the control input
  and $\vsensor_{[t]}$ is the measured output of the system.
The symbol $\oplus$ denotes concatenation and
  $\vx_{[t]} = \vstate_{[t]} \oplus \vcontrol_{[t]}$
  is the input to the EVGP model.
For example, in the case of a mechatronic system:
  $\vcontrol_{[t]}$ are the forces applied by the actuators (e.g. electric motors);
  $\vstate_{[t]}$ is the position and velocity of the joints;
  $\vsensor_{[t]}$ is the output from the sensors.

We assume independent EVGP models for each output $\vsensor_{[t]}$ in the equation
  \refeq{eq:dynamic_evgp}.
The function $g()$ in Eq. \refeq{eq:dynamic_evgp} is modeled using the
  EVGP model from Eq. \refeq{eq:evgp} and Eq. \refeq{eq:prediction}.
In the following sections we present how we can use simple Newtonian mechanics
  to define useful priors $h(\vx)^T \vbeta$ for the EVGP model.

\subsection{Priors derived from simple Newtonian dynamics}
  \label{multi-body-dynamics}

\begin{figure*}[t]
    \centering
    \subfloat[Pendulum]{
      \includegraphics[scale=0.7]{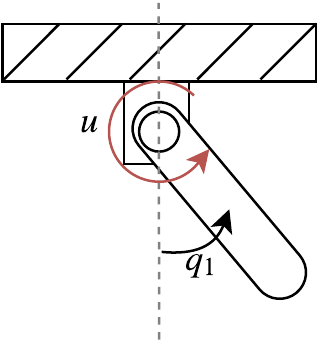}
      \label{figure:pendulum}
    } \;\;\;\;
    \subfloat[Cartpole]{
      \includegraphics[scale=0.7]{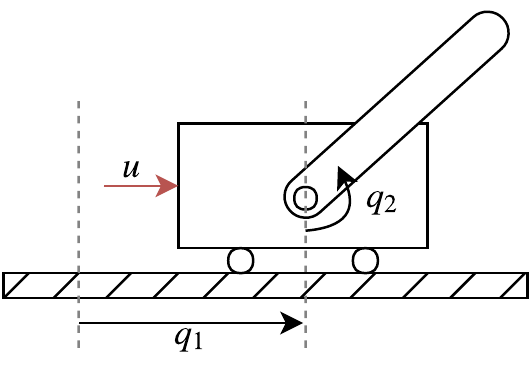}
      \label{figure:cartpole}
    } \;\;\;\;
    \subfloat[Acrobot]{
      \includegraphics[scale=0.7]{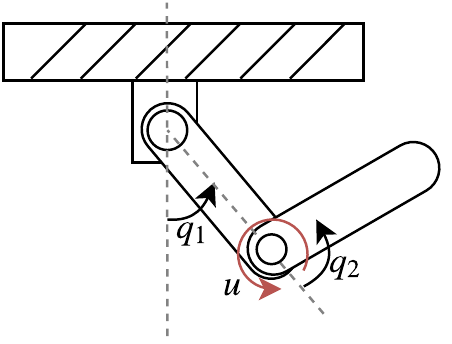}
      \label{figure:acrobot}
    } \;\;\;\;
    \subfloat[IIWA \citep{drake}]{
      \includegraphics[scale=0.25]{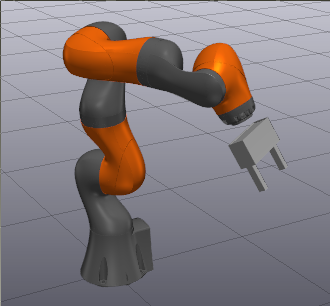}
      \label{figure:IIWA}
    } \;\;\;\;
    \caption{Diagrams of physical systems considered}
  \end{figure*}

Figure \ref{figure:pendulum} shows a simple example of a single rigid-body
  link. The simplest model that we can use for this system comes from Newton's
  second law $u = J \accel{q_1}$,
where $u$ is the torque applied to the system, $J$ is the moment of inertia, and
  $q_1$ is the angle of the pendulum.
Using Euler discretization method, we obtain the following state-space representation
  that serves as the prior $h(\vx)^T \vbeta$ for our EVGP model:
  \begin{align}
    \mat{q_1[t+1] \\ \vel{q_1}[t+1]} = &
      \mat{q_1[t] \\ \vel{q_1}[t]} +
      \Delta{t} \mat{\vel{q_1}[t] \\ \dfrac{1}{J}u[t]}
      \\ = &
      \underbrace{
        \mat{1 & \Delta{t} & 0 \\
             0 & 1         & \Delta{t}/J}}_{\vbeta \text{ prior mean } {\vmu_\beta}}
      \underbrace{
        \mat{q_1[t] \\ \vel{q_1}[t] \\ u[t]}}_{h(\vx_{[t]})}
          \label{eq:inertia_prior}
  \end{align}
We refer to this prior as IF (inertia+force).
$\Delta {t}$ is the discretization time and
$\mat{q_1[t] & \vel{q_1}[t]}^T$ is the state $\vz_{[t]}$ of the system.
The IF prior in Eq. \refeq{eq:inertia_prior} does not include gravitational effects.
Gravity pulls the link to the downward position with a force proportional to $\sin{q_1}$.
Hence, a prior that considers gravitational forces can be constructed by including
  $\sin(q_1[t])$:
\begin{align}
 \text{IFG}_{[t+1]} =
    \mat{1 & \Delta{t} & 0           & 0\\
         0 & 1         & \Delta{t}/J & -\gamma}
    \mat{q_1[t] \\ \vel{q_1}[t] \\ u[t] \\ \sin(q_1[t])}
    \label{eq:gravity_prior}
\end{align}
we call this prior IFG (inertia+force+gravity).
We purposely did not define $J$ and $\gamma$.
One of the advantages of the
  presented approach is that the algorithm can learn the parameters from data
  if they are not available.
If the user does not know the value of $J$ and $\gamma$, a prior with large
  standard deviation can be provided for these parameters (large $\Sigma_\vbeta$).
Although parameters like $J$ and $\gamma$ are easy to compute for a simple pendulum,
  for more complex systems they may be hard and tedious to obtain.
Our objective is to take advantage of domain knowledge and allow the model to fill
  in the gaps in knowledge.

For the rest of the paper, priors derived from Eq. \refeq{eq:inertia_prior}
  are referred as IF priors, while
  Eq. \refeq{eq:gravity_prior} priors are referred as IFG.
In the experiments (section \ref{section:experiments}) we compare the performance
  for both priors in order to illustrate how performance can be progressively
  improved with more detailed priors.

\subsection{Simplified priors for Acrobot, Cartpole and IIWA}
\label{section:acrobot_cartpole_priors}
In addition to the pendulum, we consider the Acrobot, Cartpole, and IIWA
  systems in our analysis.
For these systems, we consider much simpler priors than the exact dynamic models derived from multi-body dynamics.
We use the same principles shown in the previous section in order to
  get simple priors for the systems.
These rules are summarized as follows:

\begin{itemize}
  \item The position should increase proportional to the velocity by a factor of $\deltat$.
  \item The position should stay the same if the velocity is zero.
  \item The velocity should stay the same if no external forces are applied.
  \item For the IFG prior, gravity pulls the links to the downward position
    proportional to the sine of the angle w.r.t. the horizontal plane.
    Gravity has no effect when the links are completely down/up.
\end{itemize}

The objective with these priors is to demonstrate how extremely simplified priors extracted with simple physics can be used to improve performance of data-driven models. 
Figure \ref{figure:acrobot} shows a diagram of the Acrobot system.
A simple prior for this system can be constructed using the prior in Eq. \refeq{eq:inertia_prior} for each one of the links of the Acrobot:
    \begin{align}
     \text{IF} &=
        \mat{1 & 0 & \deltat & 0       & 0   \\
             0 & 1 & 0       & \deltat & 0   \\
             0 & 0 & 1       &         0 & 0   \\
             0 & 0 & 0       &         1 & \gamma_1 \\ }
        \mat{q_1[t] \\ q_2[t] \\ \vel{q_1}[t] \\ \vel{q_2}[t] \\ u[t]}
         \\
     \text{IFG} &=
        \mat{1 & 0 & \deltat & 0       & 0        & 0        & 0         \\
             0 & 1 & 0       & \deltat & 0        & 0        & 0         \\
             0 & 0 & 1       & 0       & 0        & -\gamma_2 & -\gamma_3  \\
             0 & 0 & 0       & 1       & \gamma_1 & 0        & -\gamma_4  \\ }
        \mat{q_1[t] \\ q_2[t] \\ \vel{q_1}[t] \\ \vel{q_2}[t] \\
             u[t] \\
             \sin_1  \\ \sin_{12} }
        \label{eq:acrobot_priors}
    \end{align}
  where $\sin_1 = \sin(q_1[t])$, and $\sin_{12} = \sin(q_1[t] + q_2[t])$.
In this case, the input $u[t]$ only drives the second link.
The IF and IFG priors for the Cartpole (Figure \ref{figure:cartpole})
  are the following:
\begin{align*}
   \text{IF} &=
      \mat{1 & 0 & \deltat & 0       & 0   \\
           0 & 1 & 0       & \deltat & 0   \\
           0 & 0 & 1       & 0       & \gamma_1 \\
           0 & 0 & 0       & 1       & 0   \\ }
    \mat{q_1[t] \\ q_2[t] \\ \vel{q_1}[t] \\ \vel{q_2}[t] \\ u[t]} \\
   \text{IFG} &=
      \mat{1 & 0 & \deltat & 0       & 0        & 0         \\
           0 & 1 & 0       & \deltat & 0        & 0         \\
           0 & 0 & 1       & 0       & \gamma_1 & 0         \\
           0 & 0 & 0       & 1       & 0        & -\gamma_2  \\ }
      \mat{q_1[t] \\ q_2[t] \\ \vel{q_1}[t] \\ \vel{q_2}[t] \\
           u[t] \\
           \sin(q_2[t])}
\end{align*}

\ifelsevier
\else
\begin{figure}[t]
  \centering
  \includegraphics[scale=0.45,trim={0 0.7cm 0 0},clip]{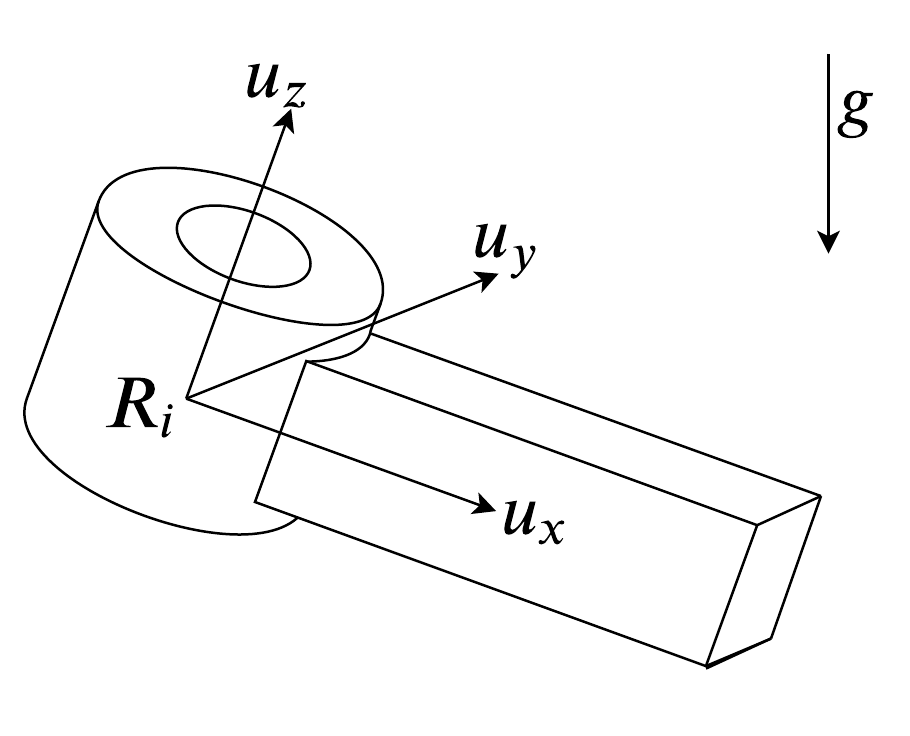}
  \caption{IFG prior for a Link in 3D}
  \label{fig:link3D}
  \end{figure}
\fi

For the IIWA system, we constructed the priors following the same rules as before, with  an approach that closely resembles to that used for the Acrobot.
The priors were constructed using the prior in Eq. \refeq{eq:inertia_prior} for each one of the IIWA links.
This results in a matrix $\vbeta$ with a similar diagonal structure than the matrices in Eq. \refeq{eq:acrobot_priors}.
To compute the IFG factors for $h()$, we derived an equation that can be used for a general 3D link (see Fig. \ref{fig:link3D}).
The equation uses forward kinematics to evaluate the contribution of the torque applied by gravity ($\vg$): 
\begin{align*}
    h^i_{IFG} \triangleq \group{\vu_x \times \vR_i^T \vg} \cdot \vu_z
\end{align*} 
where $\times$ denotes cross product, $\cdot$ denotes dot product, and 
$\vu_x$ and $\vu_z$ are unit vectors.
The matrix $\vR_i$ is the rotation matrix of the frame attached to link $i$.
This matrix is computed using classic forward kinematics, therefore its value depends on the value of $\vq$.
This equation basically computes the torque applied by gravity
to the axis of rotation $\vu_z$.
We only use unitary vectors as we assume that information like the position of the center of gravity will be captured by the posterior of $\vbeta$, i.e. the learned parameters.

These priors are extremely simple as they do not consider friction or coriolis/centrifugal
  forces.
However, they provide important information about the mechanics of the system.

\section{Experiments}\label{section:experiments}

  In order to evaluate the performance of the presented model, we performed
    experiments on a set of simulated systems:
    Pendulum, Cartpole,  Acrobot, and IIWA.
  We also performed qualitative tests on a toy-dataset to visualize the performance
    of the EVGP model.

  We used Drake \citep{drake} to simulate the Pendulum, Cartpole, Acrobot and IIWA systems
    and obtain the control/sensor data used to train and evaluate the EVGP models.
  We used the squared exponential function for the covariance kernels.
  The reason for this choice is that all the experiments involve continuous systems.
  The EVGP model was implemented using Tensorflow and the minimization of
    the negative ELBO loss was done using the ADAM optimizer \citep{kingma2014adam}.
  The experiments were run in a computer with a single GPU (Nvidia Quadro P5000)
    with an Intel(R) Xeon(R) CPU (E3-1505M at 3.00GHz).

  \subsection{Experiments on Toy Dataset}
  \label{section:toy_dataset}

\begin{figure*}[t]
  \centering
  \subfloat[GP regression]{
    \adjustimage{max size={0.35\linewidth}{0.35\paperheight}}{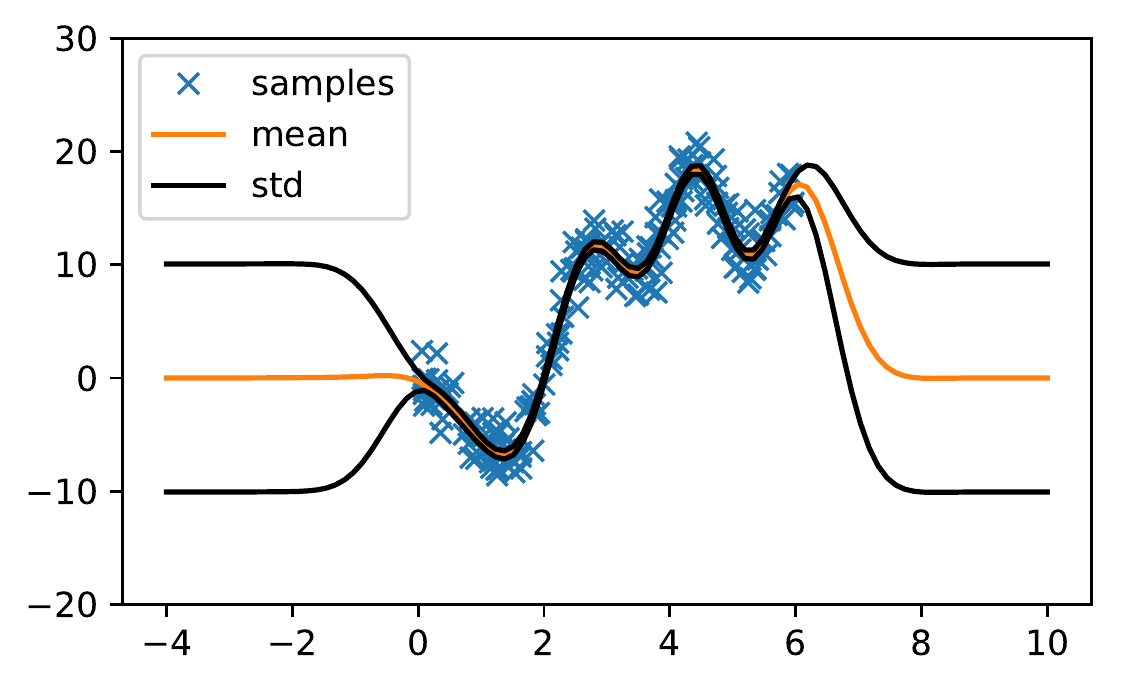}
    \label{fig:gp_estimate}
    }
  \subfloat[VGP regression (m=10)]{
    \adjustimage{max size={0.35\linewidth}{0.35\paperheight}}{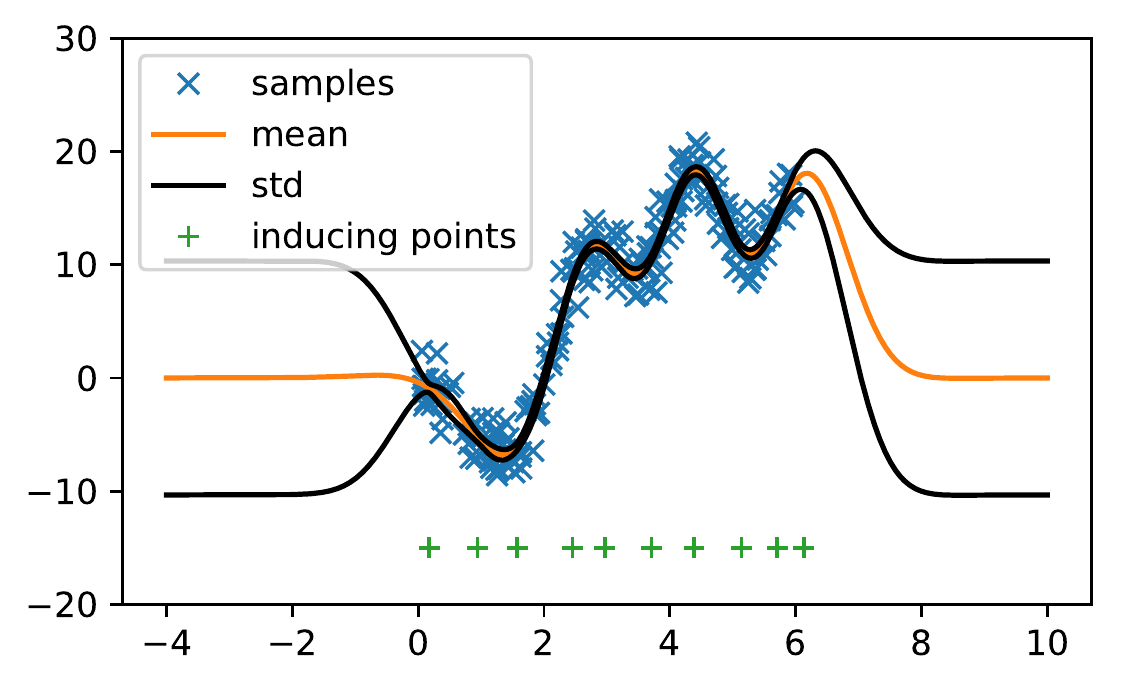}
    \label{fig:vgp_estimate}
  } \\
  \subfloat[EGP regression]{
    \adjustimage{max size={0.35\linewidth}{0.35\paperheight}}{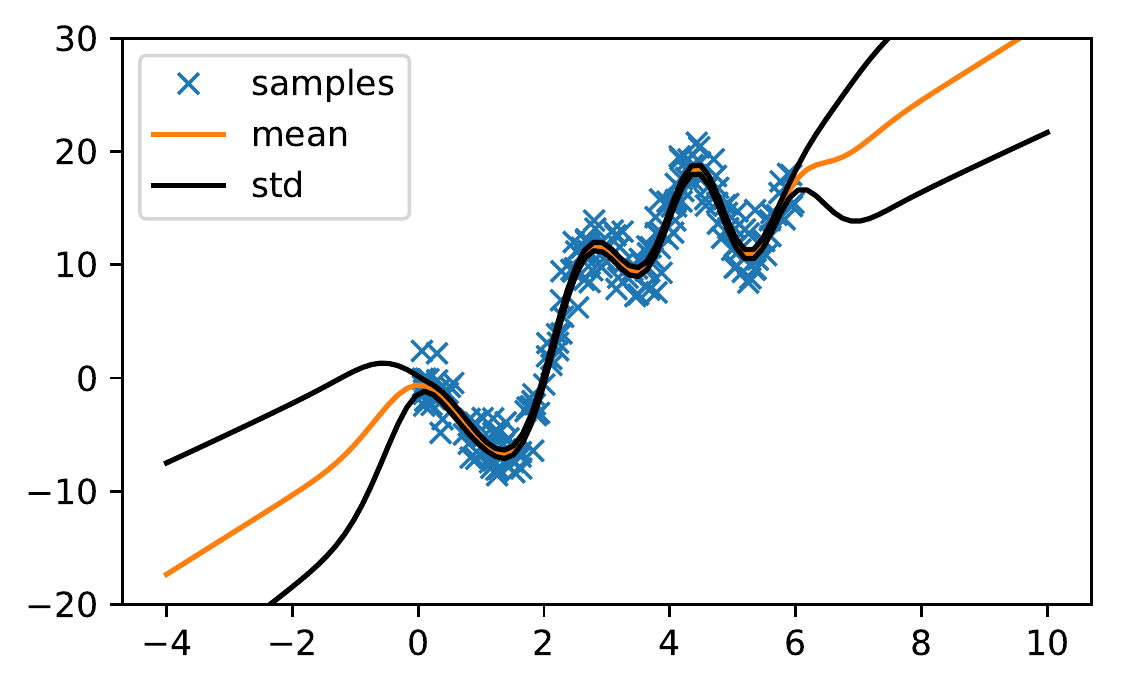}
    \label{fig:egp_estimate}
  }
  \subfloat[EVGP regression (m=10)]{
    \adjustimage{max size={0.35\linewidth}{0.35\paperheight}}{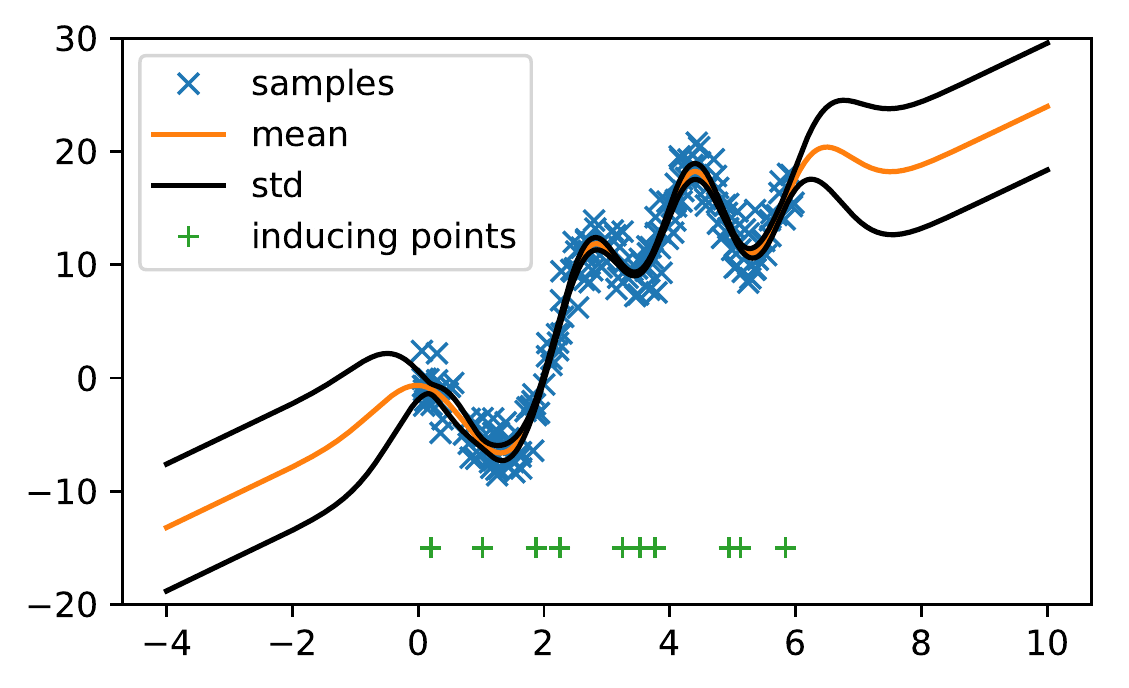}
    \label{fig:evgp_estimate}
  }
  \caption{Regression on Toy Dataset.
  The presented EVGP model provides a tool for the user to control
    the global shape of the learned function without constraining the complexity.
  The GP kernels model local non-linear behavior that the global function $h(\vx)$ is
    unable to model.}
  \label{fig:toy_dataset}
  \end{figure*}

  The toy dataset is intended to serve as an illustration of the behavior of the
    EVGP model and visualize the qualitative differences between several GP models.
  We use a modified version of the toy dataset used in
      \citep{snelson2006sparse} \citep{titsias2009variational}.
    The dataset\footnote{Obrained from http://www.gatsby.ucl.ac.uk/\textasciitilde{}snelson/SPGP\_dist.tgz}
      is modified as follows:
    \begin{align*}
      (y, x) \leftarrow ( 6 y + 3 x, x)
    \end{align*}
    The modification is intended to provide a global linear trend to the data (See Figure \ref{fig:toy_dataset}).
    Figure \ref{fig:toy_dataset} shows the distribution learned using different
      versions of a Gaussian Process.
    Figures \ref{fig:gp_estimate} and Figure \ref{fig:egp_estimate} show the
      exact posterior distributions for a GP and EGP \citep{rasmussen2004gaussian} model,
      respectively.
    Figures \ref{fig:vgp_estimate} and \ref{fig:evgp_estimate} show
      the variational approximations obtained with a VGP \citep{hensman2013gaussian}
      and EVGP model.
    The standard deviation (black line) is used to visualize the uncertainty.
    The figures show how the uncertainty grows as we move away
      from the training dataset.

    The original dataset is composed of 200 samples, Figure \ref{fig:toy_dataset}
      shows that the variational approximations are able to successfully approximate
      their exact counterparts with as few inducing points as m=10.
    The position of the inducing points are shown with green crosses.

    In this case, the prior-knowledge that we provide to the EVGP is a simple linear
      function $h(x, \vbeta) = x \beta_1 + \beta_2$.
    Figure \ref{fig:evgp_estimate} shows how we can use the prior in order to control
      the global shape of the function.
    The figure shows how the EGP and EVGP models use the prior knowledge to fit
      the global behavior of the data (linear) while using the kernels to model
      the local non-linear behavior.

\subsection{Learning system dynamics}
We evaluated the performance of the EVGP model in learning system dynamics using
  data obtained from simulations of the Pendulum, Cartpole, Acrobot and IIWA systems.
Concretely, we evaluated the accuracy of the EVGP model with IF and IFG priors
  in predicting the next state of the system given the current control inputs and state.

\textbf{Data:}
to evaluate the difference in generalization, we sampled two different datasets
  for each system: one for training and one for testing.
For the Pendulum, Cartpole, and Acrobot, 
  the datasets were sampled by simulating the system using random
  initial states $\vstate_{[0]} \sim \alpha \; \uniform{-1, 1}$
  and random control inputs $\vu_{[t]} \sim \eta   \; \normal{0, 1}$
  drawn from uniform and normal distributions, respectively.
For the IIWA, the datasets were obtained by simulating the robot going from an 
  initial position $\hat{\vq}_{[0]} \sim \uniform{-1.5, 1.5}$ to a final position 
  $\hat{\vq}_{[T]} \sim \hat{\vq}_{[0]} + \uniform{-0.8, 0.8}$. 
The IIWA was controlled by an inverse dynamics controller provided by the Drake library.
We added normal noise with a scale of $0.1*[3, 15, 3, 2, 2, 2, 0.05]$ to the output of the controller.
We had to use a controller for the IIWA in order to maintain the state
  of the robot inside a reasonable operating region.
Table \ref{table:sample_scales} shows the values of the scales ($\alpha, \eta$)
  that were used to sample the trajectories.
These values were chosen in order to cover at least the range ($-\pi, \pi$)
    for the angles on the systems.
In Table \ref{table:sample_scales}, $H$ refers to the number of sampled trajectories,
  $\abs{\dataset}$ refers to the total number of samples.
All trajectories were sampled with $\deltat=0.03 s$.

\begin{table}[t]
  \centering
  \parbox{.9\linewidth}{
    \centering
    \caption{Parameters used to collect training and testing data}
    \label{table:sample_scales}
    \scalebox{1.0}{
    \setlength\tabcolsep{1.7pt}
    \begin{tabular}{lllllll}
      \toprule
      System    &  $\alpha$                &  $\eta$  &  $H$ & $\abs{\dataset}$   \\
      \midrule
  		Pendulum  & [$\pi$, 0.5]           &   1.0    &  48  &   4800   \\
      Cartpole   & [1.0, $\pi$, 0.5, 0.5]  &   100.0  &  48  &   4800   \\
      Acrobot    & [$\pi$, 1.0, 0.5, 0.5]  &   0.5    &  93  &   9300   \\
      IIWA       &                      &       & 465  &  23250   \\  
      \bottomrule
    \end{tabular}
    }
  }
  \ifelsevier
\else
  \\
  \vspace{0.1cm}
  \fi
  \parbox{.9\linewidth}{
    \centering
    \caption{Number of inducing points and hidden units}
    \label{table:model_parameters}
    \scalebox{1.0}{
    \setlength\tabcolsep{1.7pt}
    \begin{tabular}{lllll}
      \toprule
                  &  Pendulum   &   Cartpole   &   Acrobot   &  IIWA \\
      \midrule
        VGP       &   40        &   100       &    250       &  500\\
        RES-VGP   &   40        &   100       &    250       &  500\\
        RES-DBNN  &   [15, 15]  &   [50, 50]  &    [60, 60]  &  [512, 512]\\
        EVGP-IF   &   10        &   60        &    150       &  350\\
        EVGP-IFG  &   10        &   60        &    150       &  350\\
      \bottomrule
    \end{tabular}
    }
  }
\end{table}

\textbf{Baseline:}
we compare the EVGP model with a standard VGP model \cite{hensman2013gaussian}, 
  a residual VGP (RES-VGP) \cite{deisenroth2011pilco} \cite{hensman2013gaussian}, and a
  residual Deep Bayesian Neural Network (RES-DBNN) \cite{kingma2015variational} \cite{marino2019modeling}.
The VGP model is based on \citep{hensman2013gaussian} and uses a zero mean prior.
The residual VGP and DBNN models assume the system can be approximated as
  $\vstate_{[t+1]} = \vstate_{[t]} + g_r(\vstate_{[t]} \oplus \vcontrol_{[t]})$.
Approximating residuals is a common approach used to simplify the work for
  GP and DBNN models \citep{deisenroth2011pilco} \citep{gal2016improving} \citep{marino2019modeling}.
The RES-DBNN model is trained using the local reparameterization trick presented in
  \citep{kingma2015variational} with Normal variational distributions for the weights.
For these set of experiments, we did not consider the exact GP and EGP models
  given the large number of samples in the training datasets.

\ifojies
    \begin{table}[h]
    \caption{Complexity comparison}
    \label{table:complexity}
    \centering
    \scalebox{0.9}{
    \begin{tabular}{lll}
        \toprule
                    &    Space      &  Time      \\
        \midrule
          VGP       &   $O(o m)$     &   $O(o m^3)$ \\
          DBNN      &   $O(L n^2)$   &   $O(L n^2)$   \\
          EVGP      &   $O(o m)$     &   $O(o m^3)$     \\
        \bottomrule
    \end{tabular}
    }
    \end{table}
\else
  \begin{wraptable}[7]{r}{0.25\textwidth}
      \vspace{-8mm}
      \caption{Complexity comparison}
      \label{table:complexity}
      \centering
      \scalebox{0.9}{
      \begin{tabular}{lll}
        \toprule
                    &    Space      &  Time      \\
        \midrule
          VGP       &   $O(o m)$     &   $O(o m^3)$ \\
          DBNN      &   $O(L n^2)$   &   $O(L n^2)$   \\
          EVGP      &   $O(o m)$     &   $O(o m^3)$     \\
        \bottomrule
      \end{tabular}
      }
    \end{wraptable}
\fi

Table \ref{table:model_parameters} shows the number of inducing points (m)
  used for the VGP and EVGP models.
The table also shows the number of hidden units for the DBNN model, where [15, 15]
  means a two-layer network with 15 units in each layer.
We used the LeakyRelu activation function.

Table \ref{table:complexity} shows the space and time complexity of the models
  considered in this paper.
In this table, $m$ is the number of inducing points,
  $o$ is the number of outputs,
  $L$ is the number of hidden layers,
  and $n$ is the number of hidden units in each layer.
To simplify the analysis, we assume all hidden layers of the DBNN
  have the same number of hidden units.
The table shows that the complexity of the VGP and EVGP models are governed by
  the matrix inversion $\vK_{mm}^{-1}$.
Because we assume completely independent VGP and EVGP models for each output of the system,
  their complexity also depends on the number of outputs $o$.
The complexity of the DBNN model is governed by the matrix-vector product
  between the weight matrices and the hidden activation vectors.
All models have constant space complexity w.r.t. the training dataset size
  $\abs{\dataset}$.
Furthermore, all models have linear time complexity w.r.t. $\abs{\dataset}$
  if we assume that training requires
  to visit each sample in $\dataset$ at least once.

\textbf{Metrics:}
for comparison, we used three metrics: 
  1) prediction error (Error), 
  2) predicted standard-deviation ($\|STD\|$), 
  3) containing ratios (CR).
Prediction error is computed as the difference between sampled values ($y$) and
  the expected estimated output ($\expected{}{\hat{y}^{(i)}}$).
$\|\text{STD}\|$ is computed as the magnitude of the predicted standard deviation:
\begin{align}
  \text{Error} = \dfrac{1}{\datasize} \sum_{i=1}^\datasize
      \ltwogroup{ y^{(i)} - \expected{}{\hat{y}^{(i)}} }
  \ifelsevier
\else
  \\
  \fi
  \|\text{STD}\| = \dfrac{1}{\datasize} \sum_{i=1}^\datasize
      \ltwogroup{ \func{\text{STD}}{ \hat{y}^{(i)} } }
\end{align}
where $\datasize$ is the number of samples in the respective dataset.
The expected output ($\expected{}{\hat{y}^{(i)}}$) for the EVGP model is equal
  to $\mu_{\hat{\vg} | \hat{\vx}}$ in Eq. \refeq{eq:prediction}.
For the DBNN model, the expectation is estimated using Monte-Carlo.

The containing ratios (CR) are the percentage of values covered by the estimated
  distribution $\hat{y}$.
We consider containing ratios for one, two and three standard deviations
  (CR-1, CR-2, and CR-3 respectively).
We expect the best model to have Error and $\|\text{STD}\|$ close to zero, 
  while best containing ratios will be closer to the (68-95-99.7) rule of standard distributions.

\textbf{Results:}
Table \ref{table:results} shows the prediction error and CR scores obtained in
  the testing dataset.
EVGP-IF and EVGP-IFG refers to the use of an IF or IFG prior, respectively.
We can observe a considerable improvement on the testing error and CR scores
  when using EVGP models.
EVGP models provided the lowest error and best CR scores. 
We also see a progressive improvement on the testing error when using more
  detailed priors.
The IFG prior provided lower prediction errors when compared with the IF prior.
The EVGP-IFG model provided the estimations with the lowest prediction error, 
  with low $\|STD\|$ and best CR scores.
Table \ref{table:results} also shows that the EVGP model achieved the best performance
  using fewer number of parameters.

\newcommand{\tbf}[1]{\textbf{#1}}
\newcommand{\rot}[1]{\multirow{4}{*}{\rotatebox[origin=c]{90}{#1}}}
\newcommand*\rotY{\multicolumn{1}{R{90}{0.1em}}}
\begin{table}[t]
\centering
\caption{Results on Testing Dataset}
\label{table:results}
\scalebox{0.95}{
\begin{tabular}{llrrr|>{\columncolor[gray]{0.8}}r|rr}
\toprule
   & Model &   CR-1 &   CR-2 &   CR-3 &  Error     & \rotY{$\|STD\|$} &  \rotY{\# Param.}\\
\midrule
\rot{Pendulum} 
    & VGP        & 0.582 & 0.841 & 0.910 &       0.343  & 0.273  &  407   \\
    & RES-VGP    & 0.833 & 0.951 & 0.974 &       0.155  & 0.216  &  407   \\
    & RES-DBNN   & 0.691 & 0.908 & 0.959 &       0.017  & 0.010  &  634   \\
    & \textbf{EVGP-IF}    & 0.936 & 0.986 & 0.993 &       0.013  & 0.024  &  119   \\
    & \textbf{EVGP-IFG}   & 0.871 & 0.965 & 0.985 &  \tbf{0.005} & 0.007  &  123   \\
\midrule
\rot{Cartpole}
    & VGP        & 0.505 & 0.828 & 0.901 &       0.188  & 0.185  &  2813  \\
    & RES-VGP    & 0.518 & 0.848 & 0.924 &       0.216  & 0.153  &  2813  \\
    & RES-DBNN   & 0.336 & 0.767 & 0.901 &       0.044  & 0.036  &  6008  \\
    & \textbf{EVGP-IF}    & 0.933 & 0.980 & 0.990 &       0.016  & 0.184  &  1733  \\
    & \textbf{EVGP-IFG}   & 0.935 & 0.976 & 0.986 &  \tbf{0.011} & 0.041  &  1741  \\
\midrule
\rot{Acrobot}
    & VGP        & 0.715 & 0.861 & 0.913 &       1.100  & 2.137  &  7013  \\
    & RES-VGP    & 0.705 & 0.846 & 0.904 &       0.820  & 1.428  &  7013  \\
    & RES-DBNN   & 0.606 & 0.837 & 0.899 &       0.289  & 0.378  &  8408  \\
    & \textbf{EVGP-IF}    & 0.794 & 0.908 & 0.950 &       0.151  & 0.290  &  4253  \\
    & \textbf{EVGP-IFG}   & 0.712 & 0.893 & 0.952 &  \tbf{0.131} & 0.251  &  4269  \\
\midrule
\rot{IIWA}
    & VGP        & 0.000 & 0.051 & 0.311 &       0.048  & 0.039  &  161049 \\
    & RESVGP     & 0.002 & 0.163 & 0.519 &       0.055  & 0.047  &  161049 \\
    & DBNN       & 0.001 & 0.096 & 0.346 &       0.036  & 0.025  &  143028 \\
    & \textbf{EVGP-IF}    & 0.040 & 0.353 & 0.684 &       0.037  & 0.035  &  113337 \\
    & \textbf{EVGP-IFG}   & 0.083 & 0.480 & 0.773 &  \tbf{0.027} & 0.030  &  113673 \\
\bottomrule
\end{tabular}
}
\end{table}

\begin{figure*}[t]
  \subfloat[Prediction error for all models]{
    \adjustimage{max size={0.24\linewidth}{0.4\paperheight}}{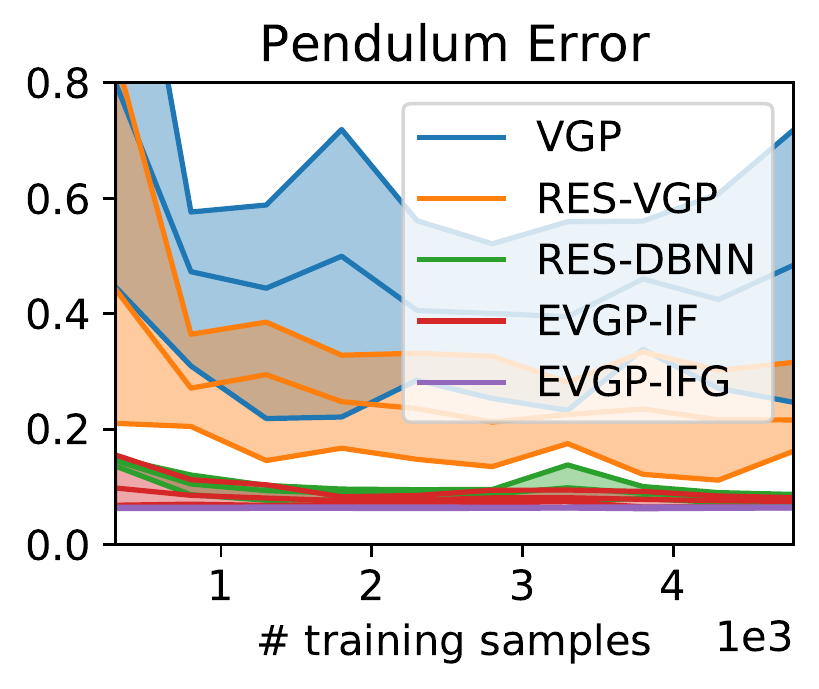}
    \adjustimage{max size={0.24\linewidth}{0.4\paperheight}}{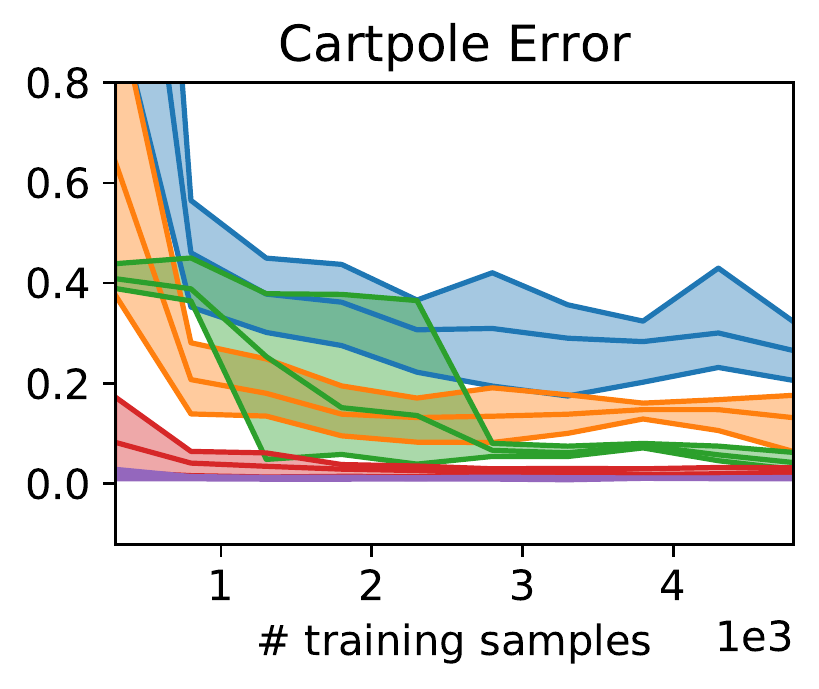}
    \adjustimage{max size={0.24\linewidth}{0.4\paperheight}}{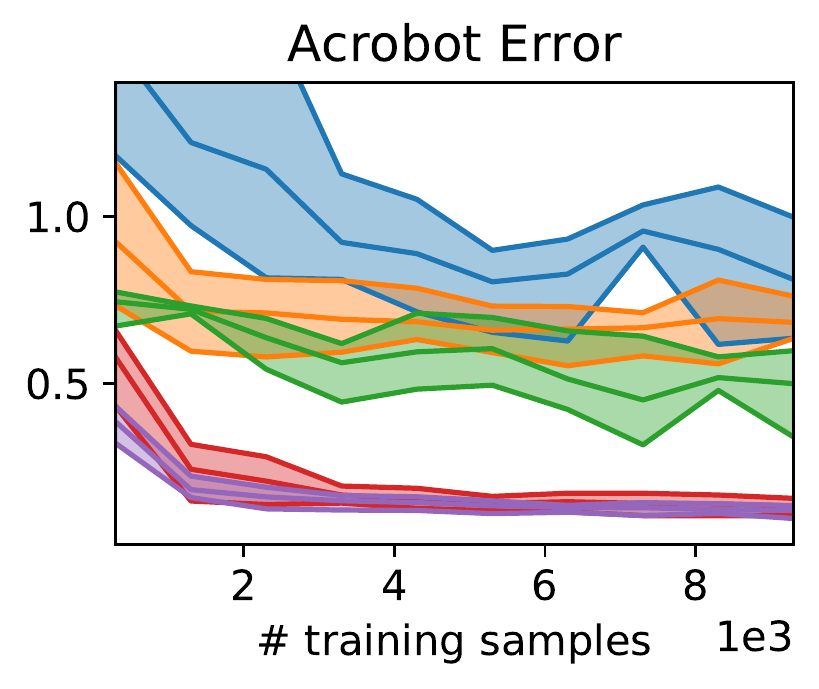}
    \adjustimage{max size={0.24\linewidth}{0.4\paperheight}}{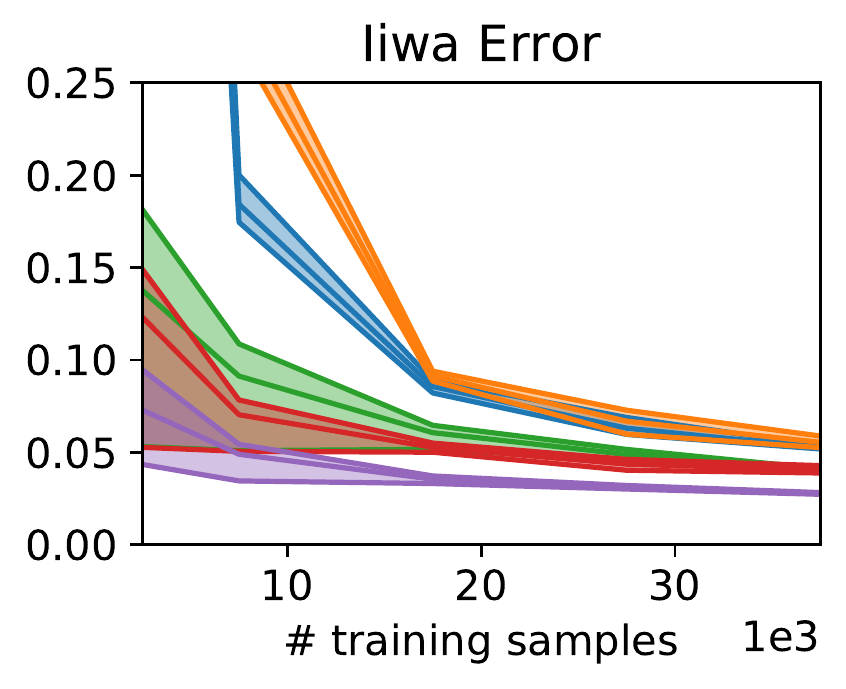}
    \label{fig:mean_norm_all}
    } \\
  \subfloat[Prediction error using IF and IFG priors]{
    \adjustimage{max size={0.24\linewidth}{0.4\paperheight}}{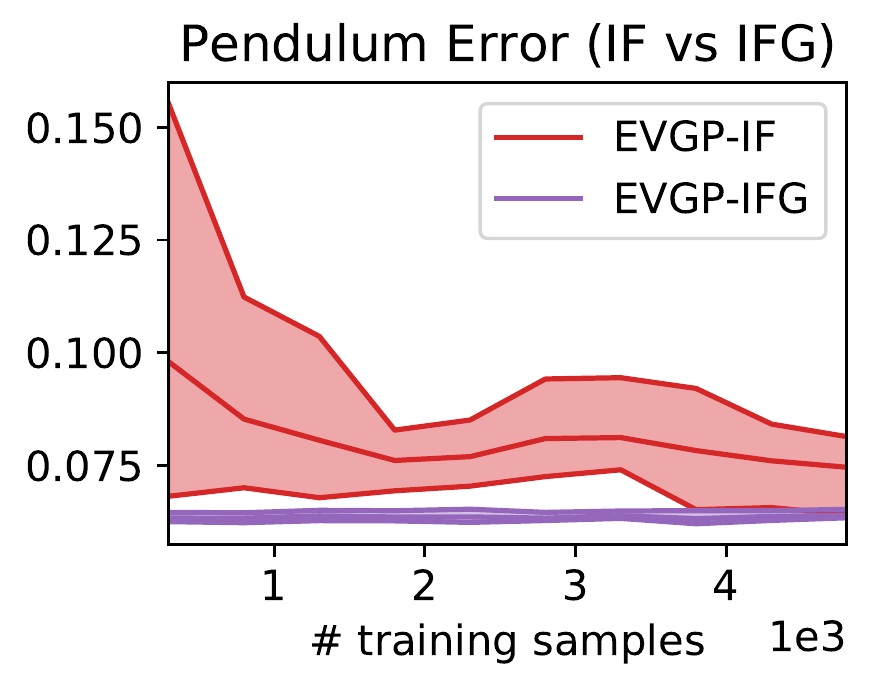}
    \adjustimage{max size={0.24\linewidth}{0.4\paperheight}}{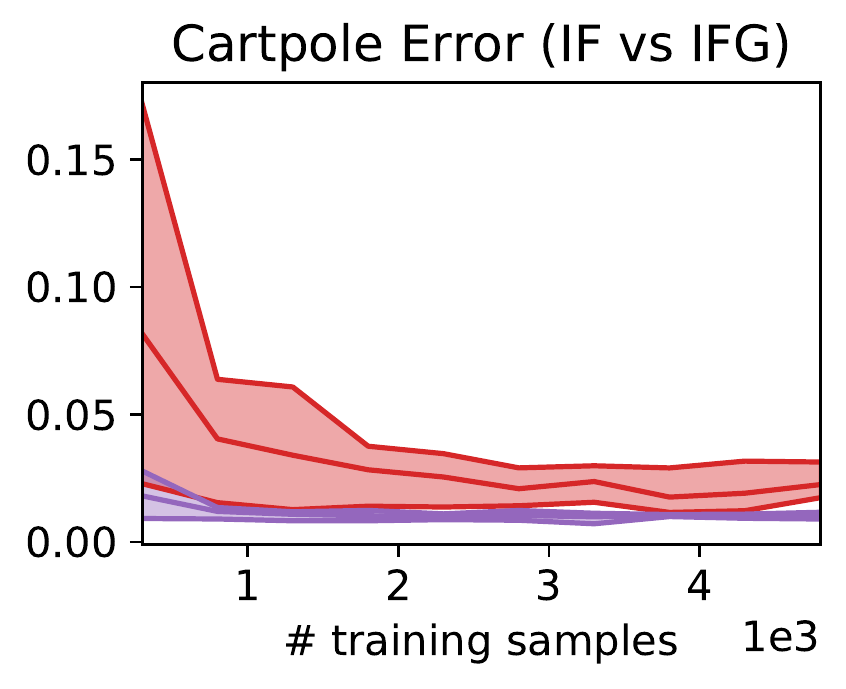}
    \adjustimage{max size={0.24\linewidth}{0.4\paperheight}}{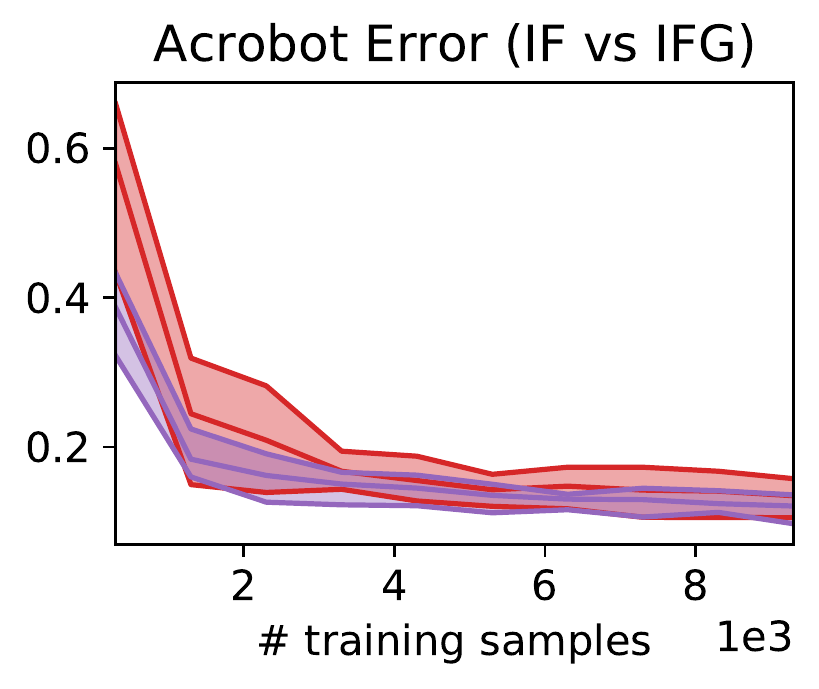}
    \adjustimage{max size={0.24\linewidth}{0.4\paperheight}}{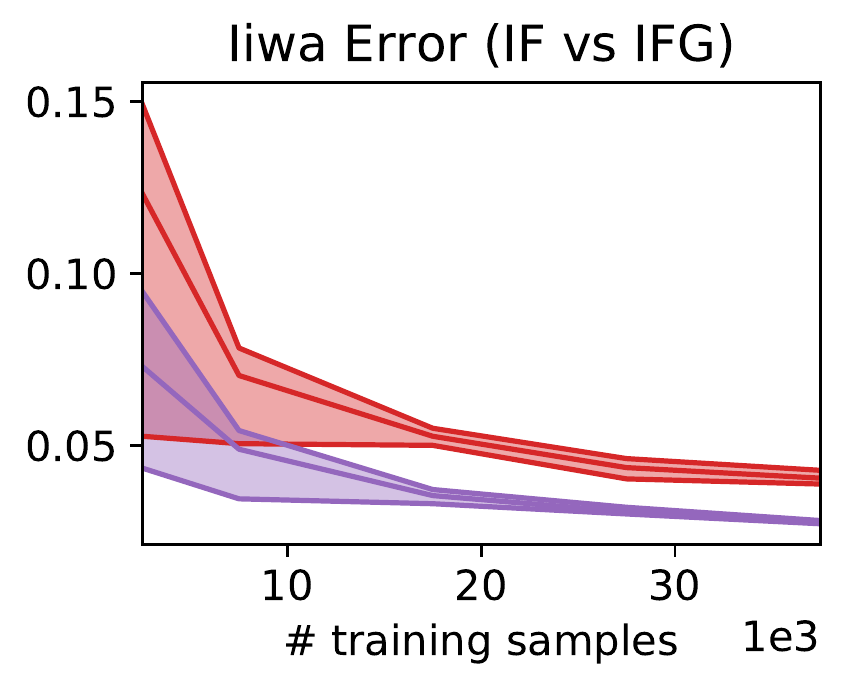}
    \label{fig:mean_norm_evgp}
    }
  \caption{Prediction error on testing dataset for increasing number of training samples.
           The EVGP model provides the most accurate predictions while requiring less
            number of samples.
           Note that the number of training samples is in the
            order of thousands.}
  \label{fig:mean_norm}
\end{figure*}

Figure \ref{fig:mean_norm} shows a comparison of the prediction error on the
  test dataset as we increase the number of training samples.
For this experiment, we kept the testing dataset fixed while samples were
  progressively aggregated into the training dataset.
The figure shows the mean, max and min values obtained for four independent runs.
Figure \ref{fig:mean_norm_all} shows a comparison that includes all models.
As expected, the prediction error is reduced as we increase the size of
  our training dataset.
The figure shows that the EVGP provides the most accurate predictions
  while requiring less number of samples.

Figure \ref{fig:mean_norm_all} also shows how the performance of VGP and RES-VGP plateaus, struggling
  to take advantage of larger datasets.
Although the RES-DBNN performs poorly with small training datasets,
  the high capacity of the RES-DBNN model allows it to take advantage
  or large datasets and improve accuracy, reducing the performance gap w.r.t.
  the EVGP models as more data is available.
Thanks to the lower computational time-cost of the RES-DBNN (see Table \ref{table:complexity}),
  this model can use a larger set of parameters without incurring in excessive training times.

Figure \ref{fig:mean_norm_evgp} shows a scaled version that only considers
  the EVGP model with different priors.
This figure shows that the IFG prior provides more accurate
  predictions when compared to the IF prior.
In the case of the pendulum, the IFG prior provides a highly accurate model
  of the system, requiring only a small number of training samples.
Figure \ref{fig:mean_norm_evgp} also shows how as training data is aggregated,
  the accuracy gap between IF and IFG priors is reduced.

The priors that we use are extremely simple,
  they are ignoring friction and coriolis/centrifugal effects.
Nonetheless, we observe a considerable
  performance improvement after providing our data-driven model basic
  information with the IF and IFG priors.

\begin{figure}[t]
    \centering
    \subfloat[$\vb$ using IF prior]{
      \includegraphics[scale=0.55]{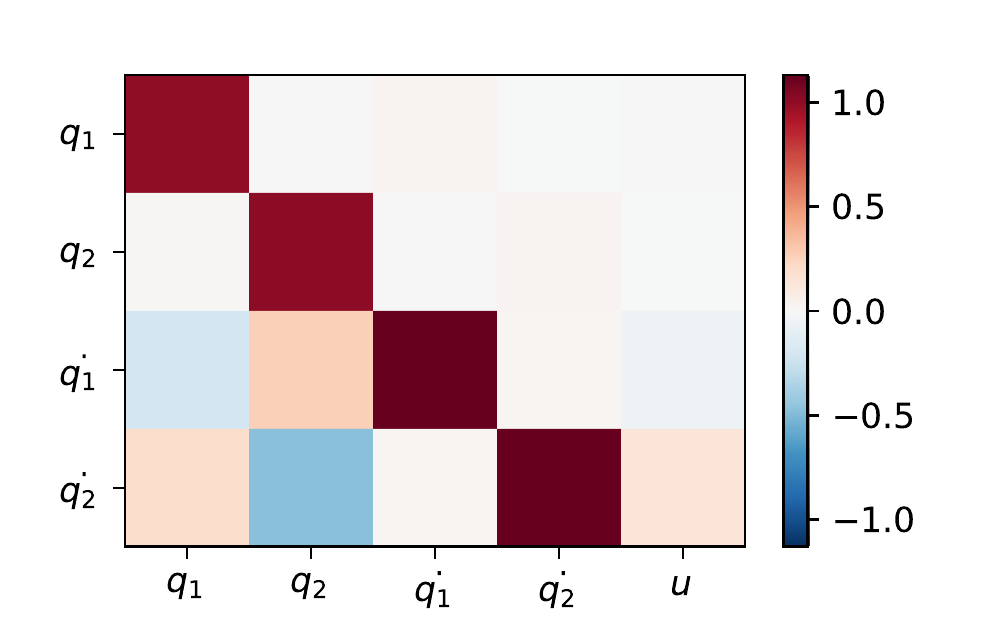}
      \label{fig:if_beta}
      } \\
    \subfloat[$\vb$ using IFG prior]{
      \includegraphics[scale=0.55]{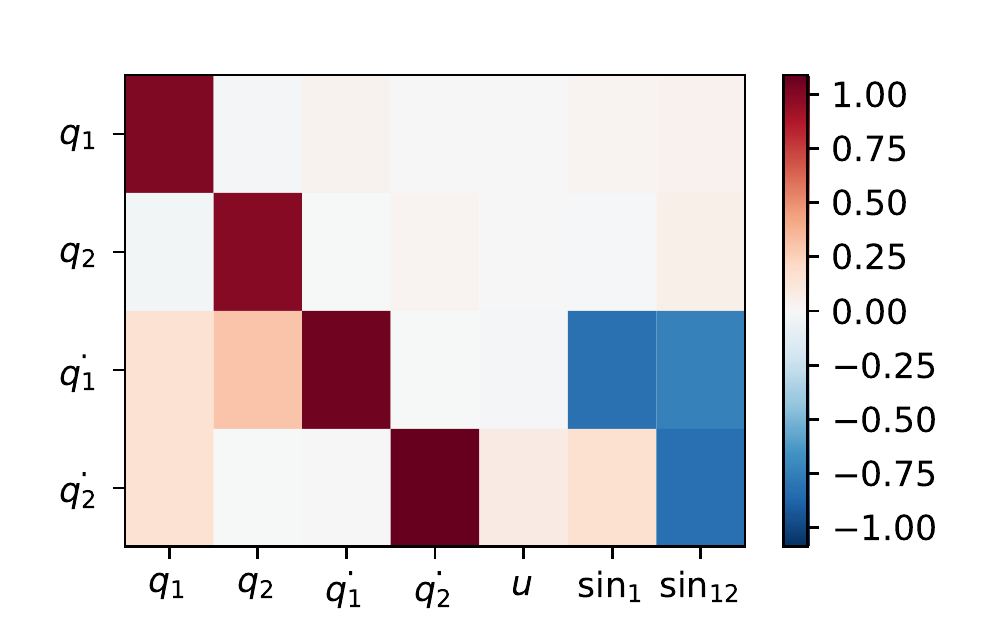}
      \label{fig:ifg_beta}
    }
    \caption{Visualization of the learned value of $\vb$ for the Acrobot}
    \label{fig:beta_acrobot}
  \end{figure}

\textbf{Understanding the learned model:}
one of the advantages of incorporating domain knowledge is that the learned model
  is easy to interpret by the domain expert.
For example, in the case of the Acrobot, the value of the parameter $\vbeta$
  can be visualized to understand and debug what the model has learned.
Figure \ref{fig:beta_acrobot} shows the value of $\vb$ learned
  with the IF (Fig. \ref{fig:if_beta}) and IFG priors (Fig. \ref{fig:ifg_beta}).

We observe in Figure \ref{fig:beta_acrobot} that the learned parameters follow a similar structure given by the prior (see Eq. \ref{eq:acrobot_priors}).
In our experiments, we did not enforce the sparse structure from the priors, i.e. zero parameters in the prior are allowed to have non-zero values in the posterior.

Figure \ref{fig:if_beta} shows that when using the IF prior, the EVGP model compensates for the missing gravity information by assigning negative values to $(\dot{q_1}, q_1)$ and $(\dot{q_2}, q_2)$.
The reason for this behavior is that $q_1 \approx \sin{q_1}$ for small $q_1$, however this approximation is no longer valid for large $q_1$.
When using IFG priors (Fig. \ref{fig:ifg_beta}), we observe that the model no longer assigns negative values to $(\dot{q_1}, q_1)$ and $(\dot{q_2}, q_2)$.
The reason is that IFG provides the values of $sin(q_1)$ and $sin(q_1 + q_2)$ which help to model the effect of gravity more accurately.

  \section{Related work}\label{section:related-work}

Incorporating prior scientific knowledge in machine-learning algorithms is an ongoing effort that has recently gained increased attention.
Hybrid modeling approaches that combine data and physics based methods are starting to gain attention in modeling cyber-physical systems \citep{rai2019driven}.
Convolutional neural networks (CNNs) have been used in modeling and simulation applications such as forecasting of sea surface temperature \citep{de2017deep} and efficient simulation of the Navier-Stokes equations \citep{tompson2017accelerating}.
Hybrid models and physics-based loss functions have been introduced in physics-informed neural networks \citep{raissi2019physics} \citep{yang2019adversarial} in order to ensure that the trained neural network satisfies physics laws.
In \citep{bikmukhametov2020combining} first principles models are combined with machine learning to create deterministic models of process engineering systems. The machine learning model (LSTM or MLP) is trained separately to predict the mismatch between a first principles model and the target solution. No training is performed on the first principles model. 
The solutions of the first principles model and the machine learning model are added to provide the final prediction.
Gaussian Processes (GP) have been used as a general purpose non-parametric model for system identification and control under uncertainty \citep{deisenroth2011pilco} \citep{bijl2017system}.
Previous work has explored using GPs to include partial model information \citep{hall2012modelling}.
In \citep{gray2018hybrid} the EGP model described in \citep{rasmussen2004gaussian} is used in combination with a simplified physics model in a thermal building modelling application.
Our work is based on the GP model with explicit features (EGP) presented in \citep{rasmussen2004gaussian}.
Variations of this model are commonly used in calibration of computer models \citep{kennedy2001bayesian} \citep{li2016integrating}.
To the best of our knowledge, we are the first to 
apply variational inference to the EGP model in order to embed simplified physics models and improve scalability.

Despite the advantages of GP models for modeling complex non-linear relations with uncertainty, GPs are computationally expensive.
A large bulk of research has focused on improving the computational requirements of GPs.
Sparse GP approximation methods are some of the most common approaches for reducing GP computation cost \citep{quinonero2005unifying}.
Bayesian approximation techniques such as Variational Inference provide a rich toolset for dealing with large quantities of data and highly complex models \citep{kingma2013auto} \citep{titsias2009variational}.
Variational approximations of a sparse GP have been explored in \citep{titsias2009variational} \citep{hensman2013gaussian}.
In \citep{frigola2014variational} a variational GP model is presented for nonlinear state-space models.
In \citep{gal2016improving} \citep{marino2019modeling}, Deep Bayesian Neural Networks (DBNNs) are proposed as an alternative to GPs in order to improve scalability in reinforcement learning problems.
Given the popularity of GP models and Variational Inference, there is an increased interest on developing automated variational techniques for these type of models \citep{nguyen2014automated} \citep{kucukelbir2017automatic}.

\section{Conclusion}
\label{section:conclusion}
In this paper, we presented the EVGP model, a stochastic data-driven model that is enhanced with domain knowledge extracted from simplified physics.
The EVGP is a variational approximation of a
  Gaussian Process which uses domain knowledge to define the mean function of
  the prior.
The priors provided a rough but simple approximation of the mechanics, informing
  the EVGP of important structure of the real system.
We compared the performance of the EVGP model against purely data-driven
  approaches and demonstrated improved accuracy and interpretability after incorporating
  priors derived from simplified Newtonian mechanics.
We showed that as we include more details into priors, the algorithm performance increases. 
We also showed how the difference between physics enhanced and purely data drive becomes smaller as we increase the availability of training data.
We demonstrated that in case of smaller training data sets, the EVGP proved its superiority over the purely data driven approaches.
Finally, we illustrate how visualizing the learned parameters can be useful to gain insights into the learned model and hence improve understandability.


\appendices

\numberwithin{equation}{section}

\section{Variational Inference}
\label{section:variational_inference}
In a Bayesian learning approach, we model the parameters $\param$ of our model
	$p(\vy \mid \vx, \param)$
  using probability distributions.
The parameters are given a prior distribution $\func{p}{\param}$ that represents our prior knowledge
  about the model before looking at the data.
Given a dataset $\dataset$, we would like to obtain the posterior distribution of our
  parameters following the Bayes rule:
\begin{align*}
  p(\param | \dataset) = \frac{\func{p}{\dataset | \param} \func{p}{\param}}
                              {\func{p}{\dataset}}
\end{align*}
Variational Inference (VI) provides a tool for approximating the posterior
  $p(\param | \dataset)$ using a variational distribution $\func{p_\phi}{\param}$
  parameterized by $\phi$.
In other words, with VI we find the value of $\phi$ such that $\func{p_\phi}{\param} \approx p(\param | \dataset)$.
The parameters $\phi$ of the distribution $\func{p_\phi}{\param}$
  are found by maximizing the Evidence Lower Bound (ELBO) between the
  approximate and real distributions:
\begin{align*}
  \phi \leftarrow \argmax_{\phi}
  \expected{\func{p_\phi}{\param}
           }{ \log \func{p}{\dataset \mid \param}}
    - D_{KL}\group{\func{p_\phi}{\param} \mid \mid \func{p}{\param}}
\end{align*}
Maximizing the ELBO is equivalent to minimizing the KL divergence
  between the variational distribution and the real distribution.

Having obtained the variational approximation $\func{p_\phi}{\param}$,
	we can approximate the predictive distribution $p(\vy | \vx, \dataset)$ as follows:
\begin{align*}
	p(\vy | \vx, \dataset) = & \expected{p(\param \mid \dataset)}{p(\vy| \vx, \param)} 
					\approx  \expected{\func{p_\phi}{\param}}{p(\vy| \vx, \param)}
\end{align*}
which is the approximated variational predictive distribution $\func{p_\phi}{\vtesty \mid \vtestX}$
	(see section \ref{section:evgp-prediction}):
\begin{align*}
	\func{p_\phi}{\vtesty \mid \vtestX} =
		\expected{p_\phi(\param)}{\func{p}{\vtesty \mid \vtestX, \param}}
\end{align*}

\section{ELBO}
\label{appendix:elbo}

Given the training dataset $\dataset = (\vy, \vX)$, the parameters $\phi$
  of $\func{p_\phi}{\param}$ are learned by minimizing the negative
  Evidence Lower Bound (ELBO). For the EVGP, the negative ELBO takes the following
  form:
\begin{equation}
\func{\loss_1}{\phi} = -\expected{\func{p_\phi}{\param}}{
     \ln \expected{g \mid \param}{\func{p}{\vy \mid \vg}} }
    +  \loss_{KL}
    \label{eq:elbo_1}
\end{equation}
where $\loss_{KL}$ denotes the KL divergence between the variational posterior and the prior
$\loss_{KL} = D_{KL}\group{\func{p_\phi}{\param} \mid \mid \func{p}{\param}}$
Note that the inner expectation in Eq. \refeq{eq:elbo_1} is taken w.r.t.
  ${g \mid \param}$, presented in Eq. \refeq{eq:g_given_w}.
Following a similar approach than \citep{hensman2013gaussian},
  we apply Jensen's inequality in the inner expectation of Eq. \refeq{eq:elbo_1}:
\begin{align*}
&\log \expected{g \mid \param}{\func{p}{\vy \mid \vg}}
  \geq \expected{g \mid \param}{\log \func{p}{\vy \mid \vg}} \\
&  \expected{g \mid \param}{\log \func{P}{\vy \mid \vg}} =
\log \normal{\vy \mid \vmu_{\vg \mid \param}, \vSigma_y}
    -\dfrac{1}{2} \Tr{\vSigma_y^{-1} \vSigma_{\vg \mid \param}}
\end{align*}
where
  $\vmu_{\vg \mid \param} = \vH_x \linw + \vmu_{\vf \mid \param}$,
  $\vSigma_{\vg \mid \param} = \vSigma_{\vf \mid \param}$.
This allows us to express the ELBO in a way that simplifies the computation
of the expectations w.r.t. the parameters $\param$.
Now, the variational loss in Eq. \refeq{eq:elbo} can be obtained by
simply computing the expectation w.r.t. the model parameters:
  \begin{align*}
  \func{\loss}{\phi}
        = & \expected{\func{p_\phi}{\param}}{
          \expected{g \mid \param}{- \log \func{p}{\vy \mid \vg}} }  +  \loss_{KL}
          \nonumber \\
        = & - \expected{\func{p_\phi}{\param}}{
            \log \normal{\vy \mid \vmu_{\vg \mid \param}, \vSigma_y}
            -\dfrac{1}{2} \Tr{\vSigma_y^{-1} \vSigma_{\vg \mid \param}}
            } \\
          & + \loss_{KL} \\
        = & - \log \normal{\vy \mid \vH_x \vb + \vK_{xm} \vK_{mm}^{-1} \va, \vSigma_y}
             \nonumber \\
          & + \dfrac{1}{2} \sgroup{
                \Tr{\vM_1 \vA}
                + \Tr{\vM_2 \vB}
                + \Tr{\vSigma_y^{-1} \vSigma_{\vf \mid \param}}
                }  \nonumber \\
          & + \loss_{KL}
\end{align*}
where
$\vM_1 = \group{\vK_{mm}^{-1} \vK_{mx}} \vSigma_y^{-1} \group{\vK_{xm} \vK_{mm}^{-1}}$,
and $\vM_2 = \vH_x^T \vSigma_y^{-1} \vH_x$.
The value of the KL-divergence is simply the sum of the divergence for both
  parameters:
  \begin{align*}
    \loss_{KL} = &
        D_{KL}\group{ \normal{\va, \vA} \mid \mid \normal{0, \vK_{mm}}} \\
         & + D_{KL}\group{ \normal{\vb, \vB} \mid\mid \normal{\vmu_{\beta}, \vSigma_{\beta}} }
  \end{align*}

\section{Mini-batch optimization}
\label{appendix:minibatches}
  In order to make the model scalable to very large datasets, the ELBO can be
    optimized using mini-batches. Following \citep{gal2016dropout}, assuming
    the samples are i.i.d., the loss for a mini-batch $(\vy, \vX)$ composed of
    $\abs{\vX}$ number of samples can be expressed as follows:
    \begin{align*}
        \func{\loss}{\phi} = - \dfrac{1}{\abs{\vX}}
          \expected{p_\phi(w)}{ \func{\ln}{ p(\vY|\vX,w)} }
              + \dfrac{1}{\abs{\dataset}} \loss_{KL}
        \end{align*}
  where $\abs{\dataset}$ is the total number of samples in the training dataset.

\ifojies

\fi

\begin{IEEEbiography}
[
{\includegraphics[width=1in,height=1.25in,trim={0.7cm 0 0.7cm 0.7cm},clip,keepaspectratio]{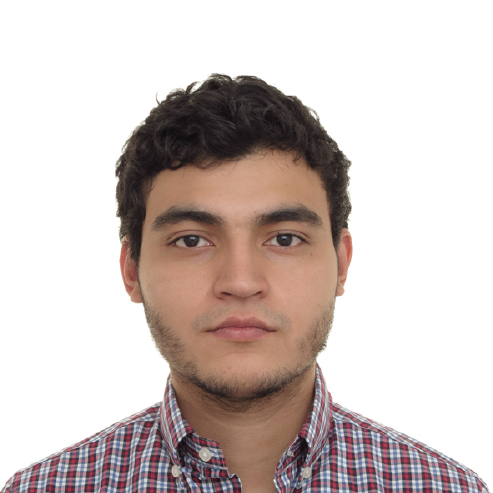}}
]
{Daniel L. Marino}
(marinodl@vcu.edu) received his B.Eng. in automation engineering from
  La Salle University, Colombia, in 2015.
He is currently a research assistant and a doctoral student at Virginia Commonwealth University.
His research interests include stochastic modeling, deep learning and optimal control.
\end{IEEEbiography}

\vskip -2\baselineskip plus -1fil

\begin{IEEEbiography}
[
{\includegraphics[width=1in,height=1.25in,clip,keepaspectratio]{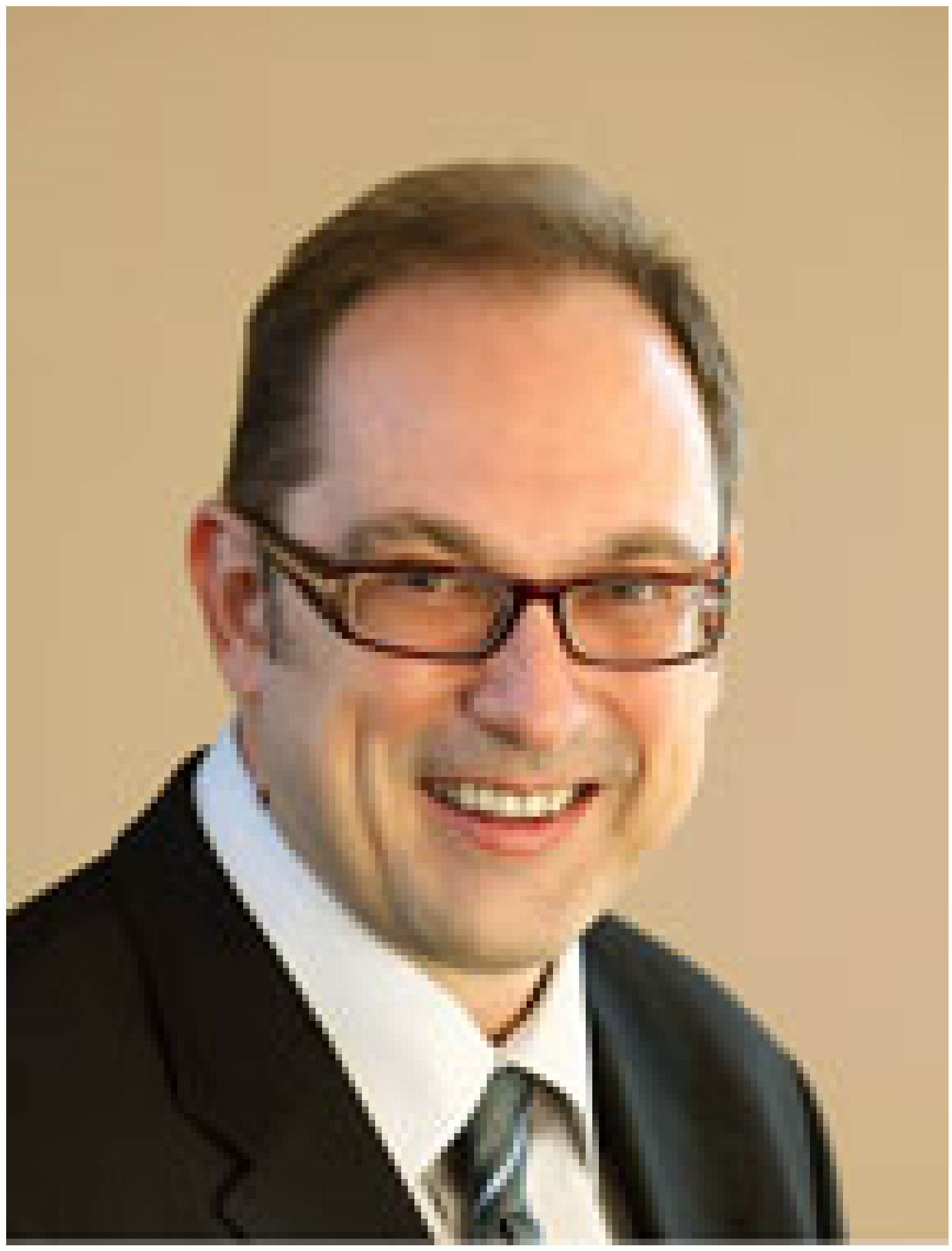}}
]
{Milos Manic} (misko@ieee.org)
 (SM'06-M'04-StM'96)
  received the Dipl.Ing. and
  M.S. degrees in electrical engineering and computer science from the
  University of Ni\v{s}, Ni\v{s}, Serbia in 1991 and 1997 respectively, and
  the Ph.D. degree in computer science from the University of Idaho in 2003.
Dr. Manic is a Professor with Computer Science Department and Director of
  VCU Cybersecurity  Center at Virginia Commonwealth University.
He completed over 30 research efforts in the area of data mining and
  machine learning applied to cybersecurity, critical infrastructure protection,
  energy security, and resilient intelligent control.
Dr. Manic has given over 30 invited talks around the world, authored over 180
  refereed articles in international journals, books, and conferences, holds
  several U.S. patents and has won 2018 R\&D 100 Award for
  Autonomic Intelligent Cyber Sensor (AICS).
He is an officer of IEEE Industrial Electronics Society, founding chair of
  IEEE IES Technical Committee on Resilience and Security in Industry,
  and general chair of IEEE IECON 2018, IEEE HSI 2019.
\end{IEEEbiography}

\EOD

\end{document}